%% file: main.tex
\pdfoutput=1
\documentclass[10pt,twocolumn,letterpaper]{article}
\input{preamble}
\begin{document}
\maketitle
\input{sections/main_content}

\input{sections/references}
\clearpage
\maketitlesupplementary
\appendix
\input{appendix/appendix}

\end{document}

%% file: preamble.tex
\usepackage[pagenumbers]{cvpr}
\usepackage[T1]{fontenc}
\usepackage[utf8]{inputenc}
\usepackage{amsmath,amssymb,mathtools}
\usepackage{booktabs,multirow,tabularx,array,longtable}
\usepackage{graphicx}
\definecolor{cvprblue}{rgb}{0.21,0.49,0.74}
\usepackage[pagebackref,breaklinks,colorlinks,allcolors=cvprblue]{hyperref}
\usepackage{url}
\usepackage{tikz}
\usetikzlibrary{arrows.meta,positioning,fit,calc}

\title{SemQAM: A Unified Generative Image Codec for Joint Semantic-Channel Coding and Modulation}
\author{%
Hao Cao$^{1}$\enspace
Jingkai Ying$^{1}$\enspace
Yinhuan Huang$^{1}$\enspace
Zhijin Qin$^{1,2}$\enspace
Yuan Shen$^{1,2}$\\
$^{1}$Tsinghua University\quad
$^{2}$State Key Laboratory of Space Network and Communications
}
\hypersetup{
  pdftitle={SemQAM: A Unified Generative Image Codec for Joint Semantic-Channel Coding and Modulation},
  pdfauthor={Hao Cao, Jingkai Ying, Yinhuan Huang, Zhijin Qin, Yuan Shen}
}
\AtBeginDocument{
\crefname{appendix}{Section}{Sections}
\Crefname{appendix}{Section}{Sections}
\crefname{subappendix}{Section}{Sections}
\Crefname{subappendix}{Section}{Sections}
}

%% file: sections/main_content.tex
\input{sections/abstract}
\input{sections/introduction}
\input{sections/preliminaries}

\input{sections/method}

\input{sections/experiments}

\input{sections/related_work}

\input{sections/conclusion}
\input{sections/statements}

%% file: sections/abstract.tex
\begin{abstract}
Conventional digital image transmission typically optimizes source coding, channel coding, and modulation separately, limiting how source representations adapt to channel impairments.
Such impairments can corrupt the latent representations recovered at the receiver and propagate through autoregressive coding, degrading reconstruction quality.
We propose \textbf{SemQAM}, a unified generative image codec that supports conventional source coding and joint semantic-channel coding and modulation within a single model.
SemQAM represents latent features using vector quantization (VQ), maps the resulting indices to QAM symbols, and updates the codeword selected by the detected index, encouraging easily confused symbols to represent similar latent features.
We also use the conditional expectation of received codewords to approximate the receiver reconstruction at the transmitter, reducing error propagation during autoregressive coding without feedback.
In the special case of noiseless transmission, SemQAM reduces to a conventional source codec called \textbf{SemQAM-C}.
Experiments show that (i) SemQAM-C achieves lower bitrates than OneDC at matched LPIPS and DISTS, with approximately \textbf{27$\times$} faster decoding.
(ii) Compared with the separated SemQAM-C system using LDPC and QAM, SemQAM further achieves BD-CBR savings of \textbf{30.85\%}, \textbf{28.06\%}, and \textbf{28.80\%} on LPIPS, DISTS, and FID, respectively, demonstrating the benefit of joint optimization.
\par
\end{abstract}

%% file: sections/introduction.tex
\section{Introduction}
\label{sec:introduction}

\begin{figure}[t]
\centering
\includegraphics[width=0.95\linewidth]{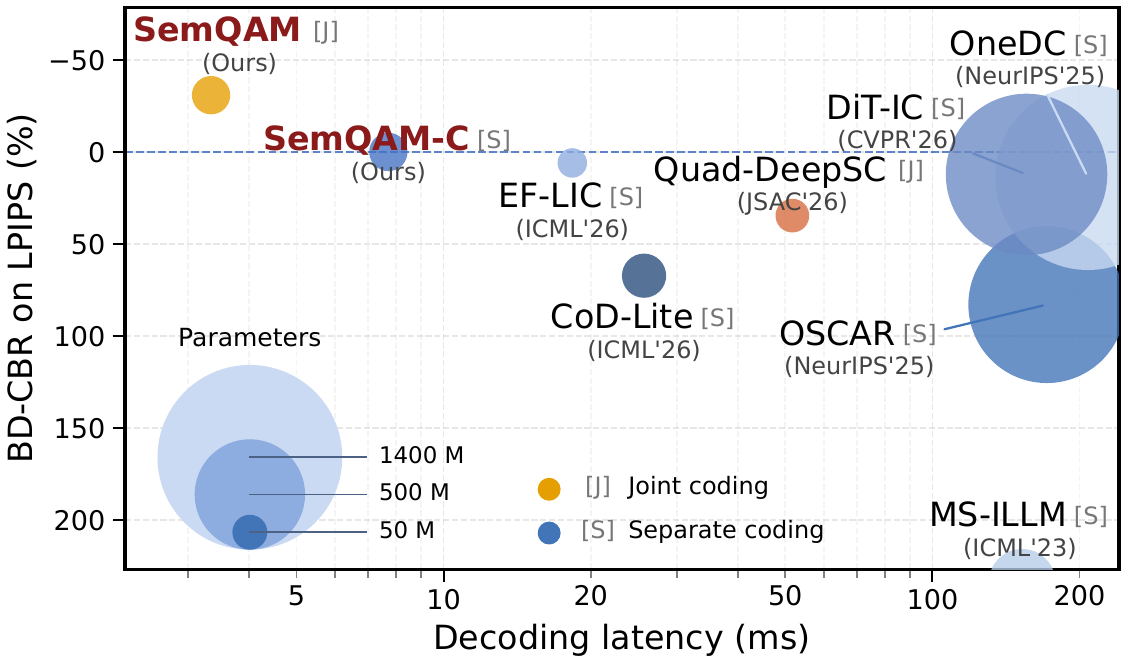}
\caption{BD-CBR versus decoding latency on Kodak at 10 dB, measured on one NVIDIA A100 GPU.
BD-CBR denotes BD-rate computed using CBR as the rate, with SemQAM-C as the anchor.
[S] denotes source codecs combined with LDPC and QAM for transmission, while [J] denotes joint coding methods that transmit learned channel symbols directly.}
\label{fig:latency_channel_use}
\vspace{-3pt}
\end{figure}

Digital image transmission aims to reconstruct images over wireless channels \citep{qin2026multimodal,xie2021deepsc}.
A conventional digital image transmission system consists of three separately optimized stages, as illustrated in \cref{fig:transmission-separated}.
(i) Source coding \citep{shannon1959coding} compresses an image into as few bits as possible at the desired reconstruction quality.
(ii) Channel coding \citep{shannon1948,gallager1962ldpc} adds redundancy to protect these bits from transmission errors.
(iii) Modulation \citep{hancock1960qam} maps the protected bits to channel symbols, such as quadrature amplitude modulation (QAM) symbols, for transmission.

However, with limited transmission resources, source coding and error protection compete for the same budget, motivating learned joint semantic-channel coding (JSCC) \citep{huang2026quaddeepsc} and joint semantic-channel coding and modulation (JSCM) \citep{ying2026token} for end-to-end optimization.
Nevertheless, existing studies largely focus on adapting source encoders to channel conditions, while the design of image codecs tailored to digital transmission remains less explored.

\begin{figure*}[t]
\centering
\begin{subfigure}{\linewidth}
\centering
\includegraphics[width=0.90\linewidth]{figures/separated_tx.pdf}
\caption{Conventional digital communication system}
\label{fig:transmission-separated}
\end{subfigure}

\begin{subfigure}{\linewidth}
\centering
\includegraphics[width=0.90\linewidth]{figures/semqam_tx.pdf}
\caption{SemQAM}
\label{fig:transmission-semqam}
\end{subfigure}
\caption{Conventional separated transmission and SemQAM. \textbf{(a)} A conventional system optimizes source coding, channel coding, and modulation separately. \textbf{(b)} Design of SemQAM. SemQAM \textit{(i)} trains VQ codewords using the detected indices at the receiver, and \textit{(ii)} uses the expectation of received codewords to approximate the receiver reconstruction during autoregressive coding.
$\mathbb{E}$, VQ, and iVQ denote conditional expectation, vector quantization, and codeword lookup, respectively.}
\label{fig:transmission}
\vspace{-3pt}
\end{figure*}

In this paper, we propose \textbf{SemQAM}, a unified generative image codec that supports conventional source coding and joint \textbf{sem}antic-channel coding and modulation using standard \textbf{QAM}, as illustrated in \cref{fig:transmission-semqam}.
Channel impairments can corrupt the latent representation recovered at the receiver, and the resulting errors can accumulate during autoregressive coding, degrading reconstruction quality.
First, SemQAM maps vector quantization (VQ) indices to QAM symbols and updates the codeword corresponding to the detected index, encouraging easily confused symbols to represent similar latent features.
Second, SemQAM computes expected received codewords from channel transition probabilities and uses them to approximate the receiver reconstruction at the transmitter, reducing error propagation during autoregressive coding without feedback.

SemQAM supports multiple signal-to-noise ratios (SNRs) and channel bandwidth ratios (CBRs) in one model, with SemQAM-C providing conventional source coding capability.
As shown in \cref{fig:latency_channel_use}, experiments on Kodak at 10 dB show that (i) SemQAM-C achieves BD-rate savings of 12.16\% and 23.59\% over the source codec OneDC~\citep{xue2025onedc} for LPIPS and DISTS, respectively.
Moreover, SemQAM-C achieves approximately 27$\times$ faster decoding while using only 4.29\% of OneDC's parameters, indicating strong source coding capability and efficiency.
(ii) Compared with the separated SemQAM-C system using LDPC~\citep{gallager1962ldpc,richardson2001ldpc} and QAM over an AWGN channel, SemQAM further achieves BD-CBR savings of 30.85\%, 28.06\%, and 28.80\% on LPIPS~\citep{zhang2018lpips}, DISTS~\citep{ding2022dists}, and FID~\citep{heusel2017fid}, respectively, demonstrating the benefit of joint optimization.
We further validate SemQAM over a Rayleigh fading channel~\citep{bourtsoulatze2019deepjscc}, where it retains its advantage over the separated SemQAM-C system.
(iii) We further evaluate SemQAM on an edge device, where it achieves faster coding speed than the compared JSCC and source coding methods, showing its potential for practical deployment.

Our contributions are summarized as follows:

\begin{itemize}
\item We introduce SemQAM, a unified generative image codec supporting both conventional source coding and joint semantic-channel coding and modulation with standard QAM across multiple SNRs and CBRs.
\item SemQAM trains VQ codewords using detected indices to incorporate channel transitions into codebook learning, and uses expected received codewords to approximate receiver reconstructions and reduce error propagation.
\item Experiments demonstrate the advantage of joint coding over separate optimization while retaining strong performance and efficiency in conventional source coding.
\end{itemize}

%% file: sections/preliminaries.tex
\section{Preliminary}
\label{sec:preliminaries}

\paragraph{Digital image transmission.}
Digital image transmission aims to reconstruct the source image through a wireless channel with high quality.
During wireless transmission, channel impairments such as noise can distort the signals and cause symbol errors, so the system must represent the image efficiently while remaining robust to these errors.

\paragraph{Source coding.}
Source coding removes redundancy from the source representation to reduce the number of bits or symbols needed for reconstruction.
Autoregressive coding reduces redundancy among latent features~\citep{balle2018hyperprior,minnen2018joint}.
In EF-LIC~\citep{cao2026eflic}, previously reconstructed groups predict an affine transform for the current latent before VQ, removing predictable information and improving representation efficiency under a fixed VQ budget.
This also makes later quantization depend on earlier reconstructions.

\paragraph{Channel coding.}
Channel coding adds redundancy to transmitted bits so that the receiver can correct errors caused by channel impairments.
Low Density Parity Check (LDPC) codes~\citep{gallager1962ldpc,richardson2001ldpc} are among the most widely used channel codes in modern communication systems~\citep{hui2018channel}.
We therefore use LDPC to construct the separated transmission baselines in our experiments.

\paragraph{Modulation.}
Modulation converts coded bits into physical signals for transmission over the wireless channel.
Quadrature amplitude modulation (QAM)~\citep{3gpp38214v1620} is one of the most widely used modulation schemes and maps coded bits to a finite set of constellation points.
An $M$-QAM constellation contains $M$ points, so each symbol can represent $\log_2 M$ bits.
For example, 1024-QAM contains 1024 constellation points and carries 10 bits per symbol, naturally matching the 1024 possible indices of a VQ codebook.

\paragraph{Wireless channel.}
A wireless channel is the physical medium through which signals are transmitted from the transmitter to the receiver.
A common channel model is
\begin{equation}
r = hs + N,
\end{equation}
where $s$ and $r$ denote the transmitted and received symbols, $h$ denotes the channel response, and $N$ denotes additive noise.
The signal-to-noise ratio (SNR) is a common measure of channel conditions and is defined as
\begin{equation}
\mathrm{SNR}
=
10\log_{10}
\frac{\mathbb E[|hs|^2]}{\mathbb E[|N|^2]}
\quad\mathrm{dB}.
\end{equation}
Under a fixed signal power, a larger SNR corresponds to weaker noise and a cleaner channel.
For digital image transmission, the channel bandwidth ratio (CBR) measures the transmission resources relative to the source image size.
For an RGB image of size $H\times W$ transmitted using $n$ channel uses, CBR is defined as
\begin{equation}
\mathrm{CBR}=\frac{n}{3HW},
\end{equation}
where a lower CBR indicates fewer transmission resources.

\paragraph{Joint semantic-channel coding and modulation (JSCM).}
Information theory shows that separate source and channel coding achieves optimal performance only under asymptotic conditions, with sufficiently long coding blocks and channel codes approaching capacity~\citep{shannon1959coding,kostina2013lossyjscc}.
These conditions are difficult to meet in practice, motivating JSCC.
Existing JSCC methods directly transmit continuous channel symbols over noisy channels~\citep{dai2022ntscc,huang2026quaddeepsc}.
However, continuous symbols are difficult to support on existing digital hardware, motivating JSCM with finite constellations.
Considering that QAM is widely used in practical digital communication systems, we therefore develop SemQAM for JSCM with QAM within a generative image codec that also supports conventional source coding for broader applications.

%% file: sections/method.tex
\section{Method}
\label{sec:method}

\subsection{System overview}
\label{sec:overview}

Conventional source codecs assume correct index recovery, but channel impairments can alter the detected indices and cause reconstruction errors to propagate through autoregressive coding.
To address these challenges, we design SemQAM by extending the autoregressive source codec EF-LIC~\citep{cao2026eflic} to joint semantic-channel coding and modulation.
The analysis transform maps an image $x$ to the main latent $y=g_a(x)$, while the hyper analysis transform extracts the hyperlatent $z=h_a(y)$~\citep{balle2018hyperprior}.
Both latents are represented using residual vector quantization (RVQ)~\citep{oord2017vqvae,kumar2023dac}, with the hyperlatent transmitted first to provide context for the main latent.
The receiver reconstructs $\hat z$ from its detected indices, while the transmitter maintains an approximation $\tilde z$ of the receiver hyperlatent, producing context features $\hat\phi=h_s(\hat z)$ and $\tilde\phi=h_s(\tilde z)$, respectively.

The main latent $y$ is divided into four quadtree groups $y_1,\ldots,y_4$, where later groups use earlier reconstructions as context \citep{cao2026eflic}.
The transmitter constructs the context from its approximated receiver reconstructions $\tilde y_{<i}$, while the receiver uses its actual reconstructions $\hat y_{<i}$.

\begin{equation}
(\tilde\mu_i,\tilde\sigma_i)=f_i(\tilde y_{<i},\tilde\phi),\quad
(\hat\mu_i,\hat\sigma_i)=f_i(\hat y_{<i},\hat\phi).
\end{equation}

The transmitter normalizes the current group as $u_i=(y_i-\tilde\mu_i)\oslash\tilde\sigma_i$, where $\oslash$ denotes elementwise division, and RVQ successively quantizes its residuals.

For each RVQ stage, the transmitter selects the nearest codeword to obtain an index $J$.
Each codebook contains 1024 codewords whose indices are mapped one to one to Gray-labeled 1024-QAM symbols~\citep{cho2002qam}, without separate channel coding between VQ and modulation.
After wireless transmission, the receiver detects $\hat J$ and reconstructs from $c_{\hat J}$, while the transmitter maintains an approximation of the receiver reconstruction because $\hat J$ is unavailable to it.
The receiver finally reconstructs the image as $\hat x=g_s(\hat y)$.

\subsection{Codebook learning under channel impairments}
\label{sec:shaping}

In a noiseless VQ system, codeword indices are only labels~\citep{oord2017vqvae}.
Permuting these labels does not affect reconstruction as long as the encoder and decoder use the same mapping.
This symmetry is broken after the indices are mapped to QAM symbols because channel impairments may change a transmitted index $J$ into a different detected index $\hat J$.
If a QAM detection error selects a codeword that is different from the transmitted one, the reconstructed latent can change substantially, degrading the image reconstruction.

SemQAM addresses this problem through receiver-side codebook learning.
For one RVQ stage, let $u$ and $\hat u$ denote the current residuals constructed from the transmitter and receiver reconstructions, respectively, and let $c_j$ and $s_j$ denote the $j$-th VQ codeword and its mapped QAM symbol.
The receiver codebook loss is
\begin{equation}
\mathcal L_{\rm cb}
=
\left\|
\operatorname{sg}(\hat u)-c_{\hat J}
\right\|^2,
\label{eq:rx_codebook_loss}
\end{equation}
where
\begin{equation}
\begin{gathered}
J=\arg\min_j\|u-c_j\|^2,\quad r=hs_J+N,\\
\hat J=\arg\min_k|r-hs_k|^2.
\end{gathered}
\end{equation}
Here, $\operatorname{sg}$ denotes stop-gradient.
The transmitted index $J$ determines the QAM symbol entering the channel, while the detected index $\hat J$ determines which VQ codeword receives the training signal.
Thus, a sample assigned to one index at the transmitter can update the codeword corresponding to another index, incorporating the transition between transmitted and detected indices into codebook learning.

Over training samples and channel realizations, the receiver codebook learning objective $\min_{\{c_k\}}\mathbb E[\mathcal L_{\rm cb}]$ can be written as
\begin{equation}
\min_{\{c_k\}}\sum_j \mathbb P(J=j)
\mathbb E\left[
\left\|\operatorname{sg}(\hat u)-c_{\hat J}\right\|^2
\mid J=j
\right].
\label{eq:expected_codebook_loss}
\end{equation}
This receiver-side objective therefore incorporates the transition from transmitted index $J$ to detected index $\hat J$ into codebook learning.
With autoregressive coding, the detected codeword also affects the residuals and context for subsequent quantization, allowing the objective to capture both direct channel-induced codeword transitions and their effects on later index assignments.
These updates reshape the codebook and index assignments according to channel impairments.
As a result, the QAM symbol distribution changes with SNR without a target distribution or shaping loss.
We further compare the learned symbol distribution with a numerically optimized information-theoretic reference in the supplementary material.

In the noiseless setting, SemQAM recovers conventional source coding through SemQAM-C.
When the channel noise disappears,
\begin{equation}
\begin{gathered}
N=0,\quad h=1
\quad\Longrightarrow\quad
r=s_J,\quad
\hat J=J,\quad
\hat u=u,\\
\mathcal L_{\rm cb}
=
\left\|
\operatorname{sg}(u)-c_J
\right\|^2.
\end{gathered}
\end{equation}
The objective therefore reduces to standard VQ learning, with no cross-index transition to organize the otherwise arbitrary codeword labels.

\begin{figure*}[t]
\centering
\includegraphics[width=0.89\linewidth]{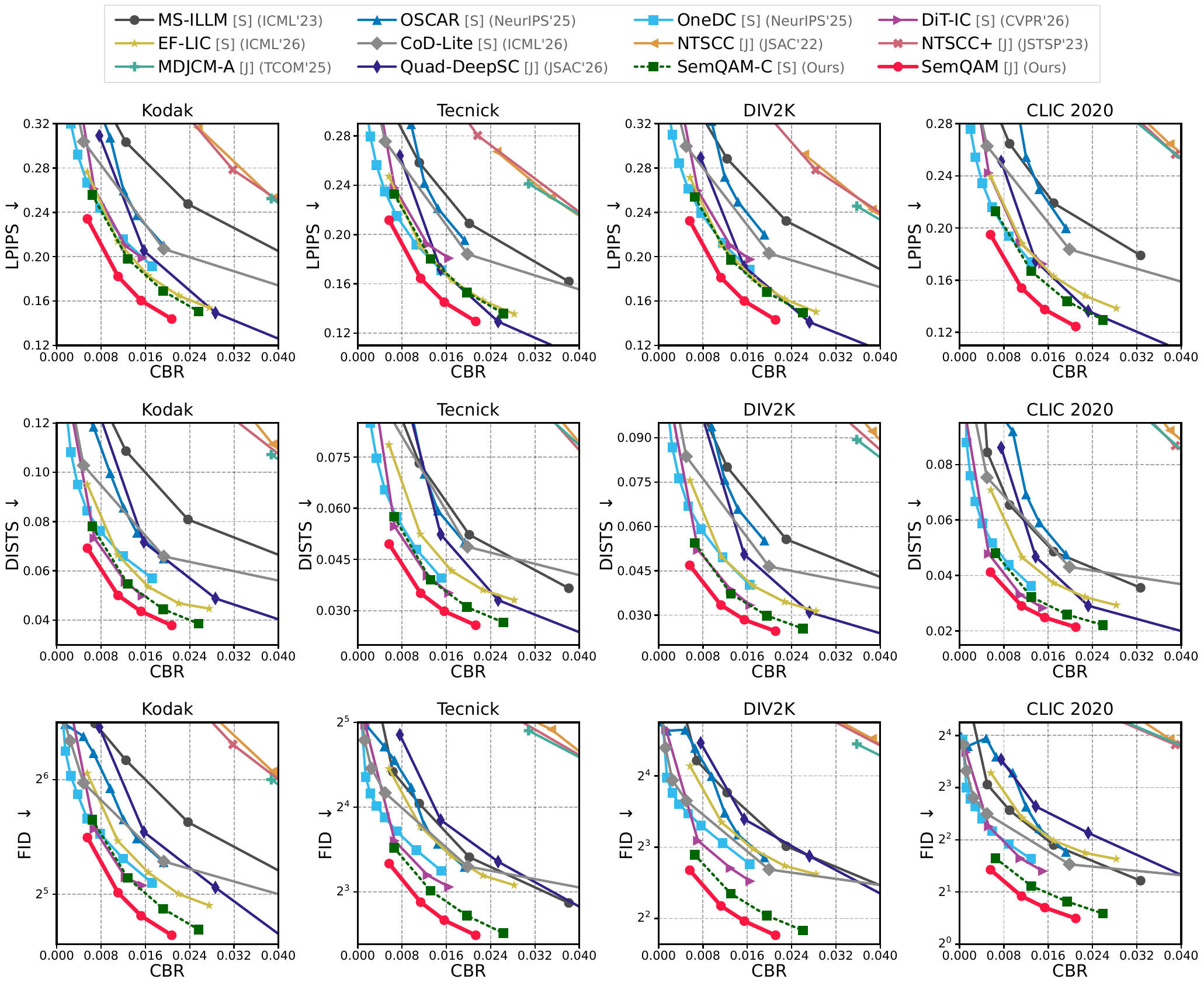}
\caption{Rate and perceptual quality comparison over a 10 dB AWGN channel.
[S] denotes source codecs combined with LDPC and QAM for transmission, while [J] denotes joint coding methods that transmit learned channel symbols directly.}
\label{fig:main_10db}
\end{figure*}

\subsection{Receiver reconstruction estimation}
\label{sec:state}

The receiver decisions described in \cref{sec:shaping} can affect the residuals and context used for subsequent quantization.
Since SemQAM uses earlier reconstructed latent groups to encode later ones, the transmitter must approximate the receiver's reconstructions.
However, it knows the sent index $J$ but not the detected index $\hat J$, so SemQAM estimates the received codeword without feedback.

For hard-decision QAM~\citep{shannon1948,farvardin1990noisyvq}, the transition matrix is
\begin{equation}
W_{kj}
=
\mathbb P(\hat J=k\mid J=j).
\end{equation}
Given a transmitted index $j$, SemQAM estimates the received codeword by its conditional mean,
\begin{equation}
\mathbb E[c_{\hat J}\mid J=j]
=
\bar c_j
=
\sum_k W_{kj}c_k.
\label{eq:expected_rx}
\end{equation}
We refer to $\bar c_j$ as the expected received codeword (Expected-RX).
As the conditional mean, $\bar c_j$ is the minimum-MSE deterministic estimate of the received codeword based only on the transmitted index.
When the channel is noiseless, $W$ becomes the identity matrix and $\bar c_j=c_j$, recovering the standard RVQ update used by SemQAM-C.

We use this estimate in transmitter-side RVQ.
For RVQ stage $l$, let $u^{(l)}$ denote the transmitter residual before quantization and $J_l$ the selected index.
The next residual is
\begin{equation}
u^{(l+1)}
=
u^{(l)}-\bar c_{J_l}.
\label{eq:tx_rvq_update}
\end{equation}
The resulting transmitter reconstructions also provide context for subsequent quadtree groups.
This expected codeword is also the commitment target,
\begin{equation}
\mathcal L_{\rm com}
=
\left\|
u^{(l)}-\operatorname{sg}(\bar c_{J_l})
\right\|^2.
\label{eq:commitment_loss}
\end{equation}
Using $\bar c_{J_l}$ in both operations keeps the transmitter residual consistent with the codeword expected to be reconstructed at the receiver.
Because later residuals and context pass through nonlinear transforms and quantizers, this remains a local estimate rather than the exact expectation of receiver reconstructions.

SemQAM also supports conventional source coding through SemQAM-C within the same model.
Under error-free transmission, the detected indices coincide with the transmitted indices, so receiver-side codebook learning reduces to standard VQ and Expected-RX recovers conventional RVQ updates.

%% file: sections/experiments.tex
\begin{table*}[t]
\centering
\caption{Results on Kodak over a 10 dB AWGN channel, with latency measured on one NVIDIA A100 GPU.
Latency is in ms, FLOPs in G, parameters in M, and BD-CBR in percent.
JSCC, Percep., and Mod. denote joint semantic-channel coding, generative optimization, and joint optimization with modulation, respectively.
SemQAM-C is the conventional source coding mode of SemQAM, evaluated with LDPC and QAM as a separated system.
SemQAM-C is the BD-CBR anchor, and lower values are better. Bold and underline mark the best and second-best values in the computation and BD-CBR columns, respectively.}
\label{tab:main_10db}
\setlength{\tabcolsep}{5.29pt}
\begin{tabular}{lrrrrrrrccc}
\toprule
& \multicolumn{4}{c}{Computation} & \multicolumn{3}{c}{BD-CBR $\downarrow$} & \multicolumn{3}{c}{Design} \\
\cmidrule(lr){2-5}\cmidrule(lr){6-8}\cmidrule(lr){9-11}
Method & Enc. & Dec. & FLOPs & Params & LPIPS & DISTS & FID & JSCC & Percep. & Mod. \\
\midrule
\input{data/main_table_rows}
\end{tabular}
\end{table*}

\begin{figure*}[t]
\centering
\includegraphics[width=0.90\linewidth]{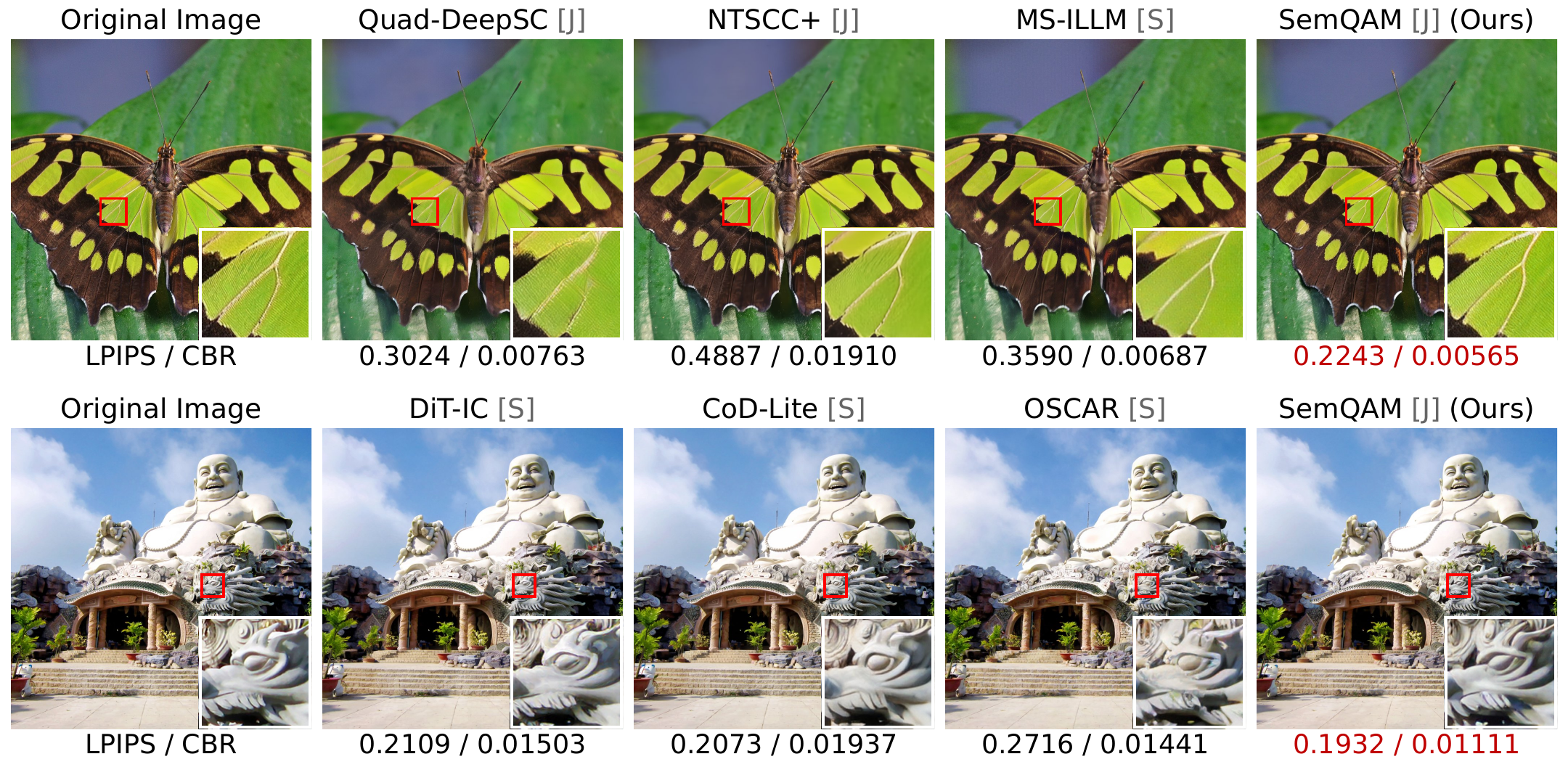}
\caption{Qualitative comparisons of SemQAM with joint transmission and separated source coding methods over a 10 dB AWGN channel.
Boxes mark local details.
Labels report LPIPS and CBR, where lower is better.}
\label{fig:visual_main}
\vspace{-10.9pt}
\end{figure*}

\section{Experiments}
\label{sec:experiments}

\subsection{Experimental setup}
\label{sec:setup}

\paragraph{Training details.}
We jointly train SemQAM at 16 operating points, covering SNR of 10, 16, and 24 dB and source coding (SemQAM-C), each with 4 CBR levels.
Following the training strategy of EF-LIC~\citep{cao2026eflic}, we train SemQAM on ImageNet~\citep{deng2009imagenet} for one week using 4 NVIDIA RTX 4090 GPUs.
SemQAM also supports other SNRs through interpolation, as detailed in the supplementary material.

\paragraph{Evaluation details.}
We evaluate SemQAM on Kodak, Tecnick, DIV2K validation, and CLIC 2020 Professional~\citep{kodak1993,asuni2013testimages,agustsson2017div2k,toderici2020clic} using LPIPS, DISTS, and FID~\citep{zhang2018lpips,ding2022dists,heusel2017fid}.
We report BD-CBR, defined as BD-rate~\citep{bjontegaard2001} computed using the channel bandwidth ratio (CBR) as the rate at matched perceptual quality.
We measure encoding and decoding latency on an NVIDIA A100 GPU and an NVIDIA Jetson Orin NX.
Our main comparisons use a 10 dB AWGN channel, following the common evaluation setting of existing JSCC and JSCM methods~\citep{dai2022ntscc,wang2023ntsccplus,zhang2024mdjcm,huang2026quaddeepsc} for direct comparison.
We also report comparisons over a Rayleigh fading channel.
Additional comparisons at other channel conditions, along with PSNR, MS-SSIM~\citep{wang2003msssim}, NIQE~\citep{mittal2013niqe}, and MUSIQ~\citep{ke2021musiq} results for SemQAM and SemQAM-C, are provided in the supplementary material.

\input{data/jetson_main_table}

\paragraph{Compared methods.}
We consider two groups of baselines for complementary comparisons.
(i) \textbf{JSCC and JSCM methods.}
We compare SemQAM with NTSCC, NTSCC+, MDJCM-A, and Quad-DeepSC~\citep{dai2022ntscc,wang2023ntsccplus,zhang2024mdjcm,huang2026quaddeepsc}.
VQ-DeepSC, DeepJSCC-Q, and CAVQ~\citep{fu2023vqdeepsc,meng2025cavq,tung2022deepjsccq} are discussed in the supplementary material.
(ii) \textbf{Source coding methods.}
We compare with MS-ILLM, OSCAR, OneDC, DiT-IC, EF-LIC, and CoD-Lite~\citep{muckley2023msillm,guo2025oscar,xue2025onedc,shi2026ditic,cao2026eflic,jia2026codlite}, combined with LDPC and QAM for separated transmission.
All methods are evaluated over 10 channel realizations using Sionna~\citep{hoydis2022sionna}.
For separated systems, failed transmissions are retransmitted, with the additional channel uses included in CBR.
We search the 3GPP modulation and coding scheme (MCS) table~\citep{3gpp38214v1620} and select the configuration achieving the lowest CBR.
These procedures optimize the transmission efficiency of separated systems, supporting a fair comparison~\citep{huang2026quaddeepsc}, detailed in the supplementary material.

\subsection{Performance comparison}
\label{sec:comparison}

\paragraph{Comparison with JSCC methods.}
Existing JSCC methods such as NTSCC, NTSCC+, and MDJCM-A are primarily optimized for PSNR, while generative JSCC remains relatively limited.
As shown in \cref{fig:main_10db,tab:main_10db}, SemQAM outperforms all four compared JSCC methods in LPIPS, DISTS, and FID, including the perceptually optimized Quad-DeepSC.
These JSCC methods directly transmit continuous-valued channel symbols without finite-constellation modulation, providing an idealized performance reference for modulation-constrained transmission~\citep{tung2022deepjsccq,bo2024jcm,ying2026token}.
In contrast, SemQAM operates with standard 1024-QAM modulation and still achieves better perceptual quality, demonstrating its effectiveness under practical digital modulation constraints.
In \cref{fig:visual_main}, SemQAM preserves the thin butterfly wing vein at lower CBR and LPIPS than the compared methods.

\paragraph{Comparison with separated systems.}
Among source codecs, SemQAM-C achieves lower CBR than OneDC at matched LPIPS, DISTS, and FID, remains competitive with DiT-IC, and outperforms lightweight codecs such as EF-LIC and CoD-Lite.
These results demonstrate that SemQAM is not limited to joint transmission and can also be applied to conventional source coding in broader scenarios.
Building on this strong baseline, SemQAM further achieves approximately 30\% BD-CBR savings over SemQAM-C across LPIPS, DISTS, and FID at 10 dB.
These gains demonstrate the benefit of jointly optimizing semantic coding, channel coding, and modulation.
The sculpture inset in \cref{fig:visual_main} further shows that SemQAM preserves its curved contours.

\paragraph{Rayleigh fading channel.}
Following common JSCC evaluation practice~\citep{dai2022ntscc,huang2026quaddeepsc}, we further evaluate SemQAM over a Rayleigh fading channel~\citep{bourtsoulatze2019deepjscc} at an average SNR of 10 dB.
Unlike AWGN, Rayleigh fading causes fluctuations in instantaneous SNR.
As shown in \cref{tab:rayleigh}, SemQAM achieves the lowest BD-CBR at matched LPIPS, with approximately 30\% savings over SemQAM-C across all four datasets.
These results demonstrate that SemQAM maintains a consistent advantage in transmission efficiency despite fluctuations in instantaneous SNR.

\input{data/rayleigh_table}

\begin{figure*}[t]
\centering
\includegraphics[width=\linewidth]{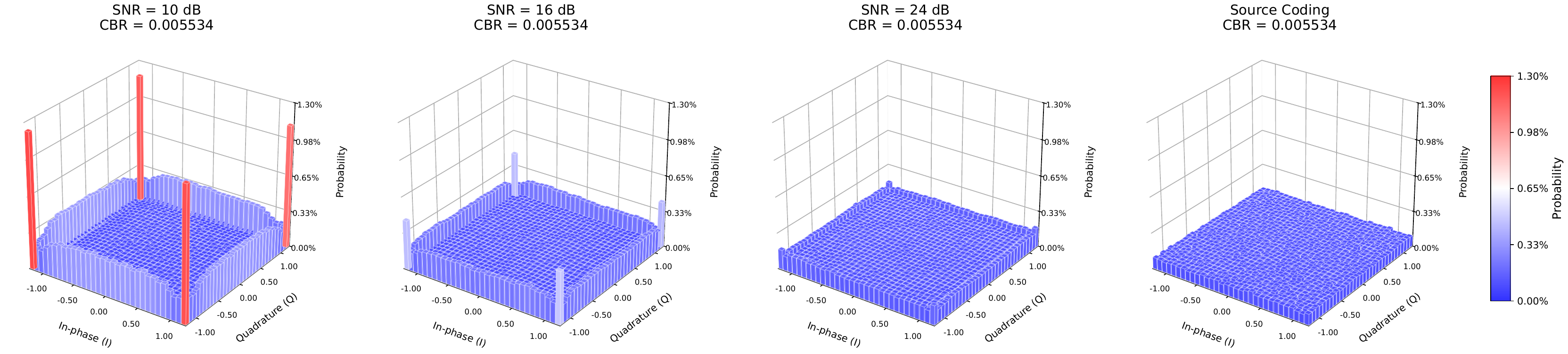}
\caption{QAM symbol usage across SNRs. Lower SNR concentrates usage on corner symbols, while higher SNR makes it more balanced.}
\label{fig:constellation_usage}
\end{figure*}

\subsection{Complexity comparison}
\label{sec:complexity}

Following \citep{cao2026grgic}, we accelerate RVQ and optimize CUDA execution to reduce coding latency.
As shown in \cref{tab:main_10db}, compared with OneDC, SemQAM-C achieves substantially faster encoding and decoding with far fewer parameters.
SemQAM also outperforms lightweight codecs such as EF-LIC and CoD-Lite in coding speed.
Although SemQAM requires more codebooks than SemQAM-C and slightly longer encoding time, it achieves faster decoding by avoiding LDPC decoding.
Additional results at higher resolutions are provided in the supplementary material.

We further evaluate SemQAM on an NVIDIA Jetson Orin NX across four resolutions, as shown in \cref{tab:jetson_main}.
SemQAM maintains the lowest encoding and decoding latency among the compared lightweight source codecs and JSCC methods, even at 4K resolution.
These results demonstrate the efficiency of SemQAM and its practical value for image transmission on edge devices.

\subsection{Constellation usage analysis}
\label{sec:usage}

As shown in \cref{fig:constellation_usage}, SemQAM favors corner symbols at 10 dB, while the distribution becomes more balanced as SNR increases and approaches uniformity in the source coding setting.
This nearly uniform distribution does not account for the different robustness of QAM symbols to channel impairments, creating a mismatch during transmission.
Through joint optimization, SemQAM implicitly learns a symbol distribution better suited to the channel without an explicit shaping objective, contributing to its gains in transmission efficiency.
We believe this alignment between source representations and the channel is a fundamental source of the gains from joint coding.

\subsection{Ablation studies}
\label{sec:ablation}

\begin{table}[t]
\centering
\setlength{\tabcolsep}{1.85pt}
\caption{Ablation configurations on Kodak at 10 dB. HP, AR, Deeper, E-RX, and RX-CB denote the hyperprior, quadtree autoregression, deeper backbone, expected received codewords, and receiver-side codebook loss, respectively.}
\label{tab:ablation}
\begin{tabular}{lccccc r r}
\toprule
\multirow{2}{*}{Model}
& \multicolumn{5}{c}{Ablation method}
& \multicolumn{1}{c}{FLOPs}
& \multicolumn{1}{c}{BD-CBR} \\
\cmidrule(lr){2-6}
\cmidrule(lr){7-7}
\cmidrule(lr){8-8}
& HP & AR & Deeper & E-RX & RX-CB
& (G) & LPIPS (\%) \\
\midrule
\input{data/ablation_table_rows}
\end{tabular}
\end{table}

To investigate the contribution of each component, we conduct the ablation study shown in \cref{tab:ablation}.
Compared with $\mathrm{M}_0$, $\mathrm{M}_1$ increases backbone capacity and improves performance, but cannot explain the full gain of SemQAM.
$\mathrm{M}_2$ adds the hyperprior to $\mathrm{M}_0$ and achieves a further improvement, demonstrating the benefit of additional context.
Compared with $\mathrm{M}_2$, $\mathrm{M}_5$ further introduces autoregression and achieves a clear gain by modeling dependencies among latent features.
Compared with $\mathrm{M}_5$, $\mathrm{M}_3$ replaces expected received codewords with transmitted codewords and performs worse, showing that Expected-RX helps reduce error propagation.
$\mathrm{M}_4$ replaces the receiver codebook loss in $\mathrm{M}_5$ with a loss based on transmitted indices, leading to a substantial degradation and demonstrating the importance of learning from receiver decisions under channel impairments.

%% file: data/main_table_rows.tex
MS-ILLM \textcolor{black!60}{(ICML'23)} & 165.53 & 152.61 & 599.52 & 181.40 & 233.82 & 324.82 & 245.39 & \textemdash & $\checkmark$ & \textemdash \\
OSCAR \textcolor{black!60}{(NeurIPS'25)} & 53.17 & 171.42 & 6485.61 & 1009.30 & 82.95 & 111.56 & 84.34 & \textemdash & $\checkmark$ & \textemdash \\
OneDC \textcolor{black!60}{(NeurIPS'25)} & 77.11 & 208.14 & 8557.18 & 1406.42 & 13.84 & 30.88 & 11.38 & \textemdash & $\checkmark$ & \textemdash \\
DiT-IC \textcolor{black!60}{(CVPR'26)} & 97.75 & 155.83 & 5743.15 & 1069.41 & 12.22 & \underline{-2.96} & \underline{-4.77} & \textemdash & $\checkmark$ & \textemdash \\
EF-LIC \textcolor{black!60}{(ICML'26)} & 17.77 & 18.34 & \textbf{279.61} & 35.74 & 5.97 & 25.72 & 37.50 & \textemdash & $\checkmark$ & \textemdash \\
CoD-Lite \textcolor{black!60}{(ICML'26)} & 9.48 & 25.71 & 771.69 & 80.15 & 67.24 & 139.64 & 86.00 & \textemdash & $\checkmark$ & \textemdash \\
\midrule
NTSCC \textcolor{black!60}{(JSAC'22)} & 36.07 & 30.12 & 533.55 & \textbf{31.36} & 359.10 & 696.09 & 494.21 & $\checkmark$ & \textemdash & \textemdash \\
NTSCC+ \textcolor{black!60}{(JSTSP'23)} & 49.77 & 49.23 & \underline{357.06} & 58.59 & 336.34 & 603.25 & 442.60 & $\checkmark$ & \textemdash & \textemdash \\
MDJCM-A \textcolor{black!60}{(TCOM'25)} & 33.48 & 59.47 & 534.96 & \underline{32.93} & 336.90 & 634.77 & 434.92 & $\checkmark$ & \textemdash & \textemdash \\
Quad-DeepSC \textcolor{black!60}{(JSAC'26)} & 28.79 & 51.71 & 531.81 & 46.65 & 34.60 & 89.99 & 98.34 & $\checkmark$ & $\checkmark$ & \textemdash \\
\midrule
SemQAM-C & \textbf{3.82} & \underline{7.71} & 381.24 & 60.34 & \underline{0.00} & 0.00 & 0.00 & \textemdash & $\checkmark$ & \textemdash \\
SemQAM & \underline{4.97} & \textbf{3.34} & 381.34 & 60.34 & \textbf{-30.85} & \textbf{-28.06} & \textbf{-28.80} & $\checkmark$ & $\checkmark$ & $\checkmark$ \\
\bottomrule

%% file: data/jetson_main_table.tex
\begin{table*}[t]
\centering
\caption{Encoding and decoding latency (ms) of SemQAM and lightweight source codecs and JSCC methods on an NVIDIA Jetson Orin NX. Bold and underline mark the best and second-best results, respectively. OOM denotes out of memory.}
\label{tab:jetson_main}
\setlength{\tabcolsep}{7.67pt}
\begin{tabular}{lrrrrrrrr}
\toprule
\multirow{2}{*}{Method} & \multicolumn{2}{c}{$512\times768$} & \multicolumn{2}{c}{1080p} & \multicolumn{2}{c}{2K} & \multicolumn{2}{c}{4K} \\
\cmidrule(lr){2-3}\cmidrule(lr){4-5}\cmidrule(lr){6-7}\cmidrule(lr){8-9}
 & Enc. & Dec. & Enc. & Dec. & Enc. & Dec. & Enc. & Dec. \\
\midrule
EF-LIC \textcolor{black!60}{(ICML'26)} & \underline{73.33} & \underline{255.88} & \underline{358.00} & \underline{1086.40} & \underline{643.16} & \underline{1952.52} & \underline{1438.57} & \underline{4353.73} \\
CoD-Lite \textcolor{black!60}{(ICML'26)} & 119.35 & 274.29 & 567.78 & 1436.33 & 975.41 & 2630.42 & 2198.84 & 15115.62 \\
Quad-DeepSC \textcolor{black!60}{(JSAC'26)} & 273.24 & 435.39 & 1512.40 & 2311.64 & 2781.69 & 4143.24 & OOM & \textemdash \\
\midrule
SemQAM & \textbf{31.10} & \textbf{42.09} & \textbf{134.57} & \textbf{191.25} & \textbf{233.85} & \textbf{332.32} & \textbf{511.68} & \textbf{735.45} \\
\bottomrule
\end{tabular}
\end{table*}

%% file: data/rayleigh_table.tex
\begin{table}[t]
\centering
\caption{Comparison of channel use at matched LPIPS over a 10 dB Rayleigh fading channel.
BD-CBR denotes BD-rate computed using CBR as the rate, with SemQAM-C as the anchor.}
\label{tab:rayleigh}
\setlength{\tabcolsep}{3.43pt}
\begin{tabular}{lrrrr}
\toprule
\multirow{2}{*}{Method} & \multicolumn{4}{c}{BD-CBR $\downarrow$} \\
\cmidrule(lr){2-5}
 & Kodak & Tecnick & DIV2K & CLIC \\
\midrule
OneDC \textcolor{black!60}{(NeurIPS'25)} & 19.98 & \underline{-10.08} & 4.31 & 4.94 \\
DiT-IC \textcolor{black!60}{(CVPR'26)} & 18.88 & 10.74 & 16.84 & 20.73 \\
EF-LIC \textcolor{black!60}{(ICML'26)} & 5.88 & 1.31 & 2.40 & 7.88 \\
CoD-Lite \textcolor{black!60}{(ICML'26)} & 67.71 & 56.25 & 61.89 & 105.38 \\
\midrule
SemQAM-C & \underline{0.00} & 0.00 & \underline{0.00} & \underline{0.00} \\
SemQAM & \textbf{-34.28} & \textbf{-33.63} & \textbf{-35.07} & \textbf{-26.19} \\
\bottomrule
\end{tabular}
\end{table}

%% file: data/ablation_table_rows.tex
$\mathrm{M}_0$ & \textemdash & \textemdash & \textemdash & $\checkmark$ & $\checkmark$ & 310.29 & 0.00 \\
$\mathrm{M}_1$ & \textemdash & \textemdash & $\checkmark$ & $\checkmark$ & $\checkmark$ & 397.52 & $-4.84$ \\
$\mathrm{M}_2$ & $\checkmark$ & \textemdash & \textemdash & $\checkmark$ & $\checkmark$ & 368.66 & $-5.15$ \\
$\mathrm{M}_3$ & $\checkmark$ & $\checkmark$ & \textemdash & \textemdash & $\checkmark$ & 381.44 & $-7.99$ \\
$\mathrm{M}_4$ & $\checkmark$ & $\checkmark$ & \textemdash & $\checkmark$ & \textemdash & 381.44 & $>100$ \\
$\mathrm{M}_5$ & $\checkmark$ & $\checkmark$ & \textemdash & $\checkmark$ & $\checkmark$ & 381.44 & \textbf{-12.29} \\
\bottomrule

%% file: sections/related_work.tex
\section{Related work}
\label{sec:related_work}

\paragraph{Learned image JSCC and JSCM.}
Learned JSCC adapts image representations to channel impairments to improve reconstruction.
DeepJSCC and NTSCC achieve this by directly transmitting continuous channel symbols~\citep{bourtsoulatze2019deepjscc,dai2022ntscc}.
NTSCC+ and Quad-DeepSC further improve performance but require separate reliable transmission of auxiliary information~\citep{wang2023ntsccplus,huang2026quaddeepsc}.
However, continuous signals are difficult to support on existing digital communication hardware, motivating JSCM.
DeepJSCC-Q uses finite constellations, while VQ-DeepSC quantizes image features for digital transmission~\citep{tung2022deepjsccq,fu2023vqdeepsc}.
JCM further learns symbol probabilities to maximize the information about source data, while MDJCM adapts to channel conditions using different modulation orders~\citep{bo2024jcm,zhang2024mdjcm}.
SemQAM enables end-to-end joint transmission and adapts symbol probabilities across channel conditions.
It maintains strong perceptual quality and transmission efficiency, while also supporting conventional source coding within the same codec for wider use.

\paragraph{Vector quantization for image transmission.}
Classical quantizers account for channel impairments in index assignment and codebook learning~\citep{farvardin1987quantizer,farvardin1990noisyvq}.
CAVQ incorporates channel transitions into learned vector quantization but does not model autoregressive dependencies or support conventional source coding~\citep{meng2025cavq}.
EF-LIC provides efficient autoregressive coding with RVQ~\citep{cao2026eflic}, assuming identical transmitter and receiver reconstructions.
SemQAM extends autoregressive image coding to noisy channels, achieving high perceptual quality and transmission efficiency while retaining conventional source coding in the same codec.

%% file: sections/conclusion.tex
\section{Conclusion}
\label{sec:conclusion}

We presented SemQAM, a unified generative image codec supporting conventional source coding and joint semantic-channel coding and modulation.
Receiver-side codebook learning implicitly adapts QAM symbol usage to channel impairments, while expected received codewords reduce error propagation during autoregressive coding without feedback.
Experiments demonstrate competitive perceptual compression and fast inference with SemQAM-C, alongside substantial transmission efficiency gains from joint optimization with SemQAM.

%% file: sections/statements.tex
\section*{LLM Usage Statement}
OpenAI ChatGPT and Codex were used to assist with language editing, checks of references, and LaTeX formatting throughout the main text and supplementary materials.
The authors are responsible for the scientific content and final manuscript.

%% file: sections/references.tex
\begingroup
\small
\bibliographystyle{ieeenat_fullname}
\bibliography{references_base,references_extra}
\endgroup

%% file: appendix/appendix.tex
\section*{Contents}
In the supplementary material, we provide the following:
\begin{itemize}
    \item \Cref{app:constellation_probability} analyzes the QAM symbol probabilities learned by SemQAM and demonstrates their similarity to a numerical reference derived from information theory.
    
    \item \Cref{app:additional_results} reports encoding and decoding latency and GPU memory at 1080p, 2K, and 4K resolutions.
    It also provides PSNR, MS-SSIM~\citep{wang2003msssim}, NIQE~\citep{mittal2013niqe}, and MUSIQ~\citep{ke2021musiq} results, comparisons at 16 and 24 dB, and validation of SNR interpolation without retraining.
    Further experiments demonstrate performance across four datasets and robustness to different initializations and channel noise realizations.
    
    \item \Cref{app:method_details} provides architectural details, analyzes codebook learning, and explains the choice of 1024-QAM.

    \item \Cref{app:protocol} explains the selection of baselines, including DeepJSCC-Q~\citep{tung2022deepjsccq}, VQ-DeepSC~\citep{fu2023vqdeepsc}, and CAVQ~\citep{meng2025cavq}, which lack available implementations.
    It also establishes a fair comparison with separated systems optimized for transmission efficiency, justifies the evaluation settings, and details the choice and computation of quality metrics.

    \item \Cref{app:visual_comparison} provides additional visual comparisons with existing methods.
\end{itemize}

\input{appendix/1_constellation_probability_analysis}

\input{appendix/2_additional_experimental_results}

\input{appendix/3_additional_method_details}

\input{appendix/4_training_evaluation_protocol}

\input{appendix/5_visual_comparison}

%% file: appendix/1_constellation_probability_analysis.tex
\section{QAM symbol probabilities from information theory}
\label{app:constellation_probability}

We compare the symbol probabilities learned by SemQAM with an information-theoretic reference obtained by maximizing the mutual information preserved after QAM detection.
This reference follows the same SNR normalization as SemQAM but considers the channel independently of image reconstruction.

\subsection{Mutual information reference}

Let $p_j=\Pr(J=j)$ denote the probability of transmitting QAM symbol $s_j$, and let $p$ denote the complete symbol distribution.
The average transmitted symbol power is
\begin{equation}
P_s(p)=\sum_j p_j|s_j|^2.
\end{equation}
For a target linear SNR $\gamma$, the corresponding noise power is
\begin{equation}
\sigma_N^2(p)=\frac{P_s(p)}{\gamma}.
\label{eq:ref_noise_power}
\end{equation}
Therefore, assigning more probability to higher-energy symbols also increases the noise power, rather than providing a higher operating SNR.

Under hard-decision QAM, the probability of detecting index $k$ when index $j$ is transmitted is
\begin{equation}
W_{kj}(p)
=
\Pr\left(
\hat J=k
\mid J=j,\sigma_N^2(p)
\right).
\label{eq:ref_transition}
\end{equation}
The transition matrix depends on $p$ because the noise power changes with the input distribution.

We consider the distribution that maximizes the mutual information between the transmitted and detected indices,
\begin{equation}
p_\gamma^*=\arg\max_p I_\gamma(J;\hat J),\quad p_j\geq 0,\quad \sum_j p_j=1,
\label{eq:ref_mi_optimization}
\end{equation}
where
\begin{equation}
I_\gamma(J;\hat J)
=
\sum_j p_j\sum_k W_{kj}(p)
\log_2
\frac{W_{kj}(p)}
{\sum_m p_m W_{km}(p)}.
\label{eq:ref_mutual_information}
\end{equation}
Unlike a fixed discrete channel, the transition matrix here depends on the distribution being optimized.
We solve this problem numerically over the probability simplex and use the result as a channel reference rather than a training target for SemQAM.

\begin{figure*}[t]
    \centering
    \includegraphics[width=\linewidth]{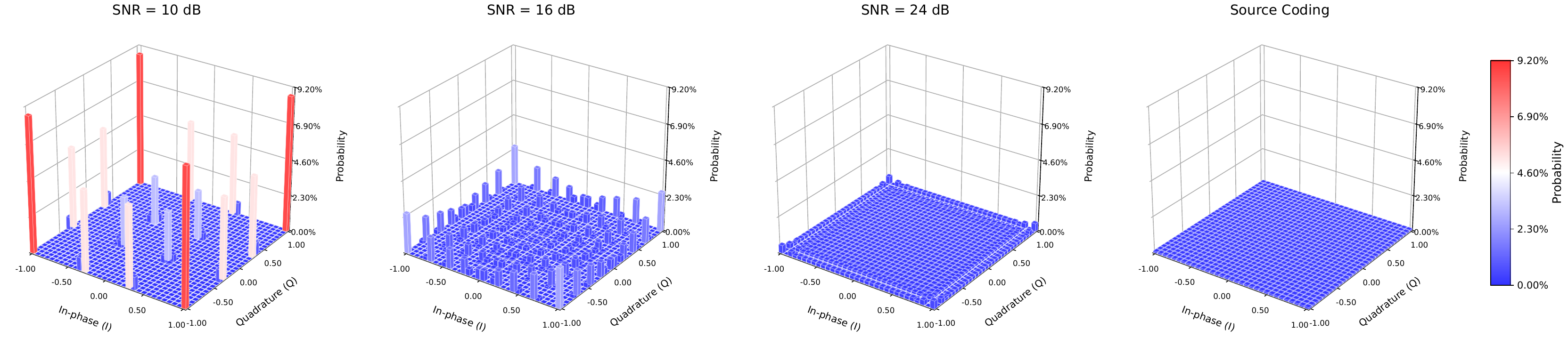}
    \caption{
    Numerically optimized 1024-QAM symbol probabilities under the average-power SNR normalization.
    Lower SNR favors corner and boundary symbols, while higher SNR produces broader symbol usage.
    }
    \label{fig:information_theoretic_distribution}
\end{figure*}

\input{data/resolution_table}

As shown in \cref{fig:information_theoretic_distribution}, the reference distribution concentrates on corner and boundary symbols at low SNR.
These symbols have fewer neighboring decision regions and are therefore less susceptible to confusion under strong noise.
The distribution balances this advantage against the increased noise power and the need to preserve information across distinct symbols.
As SNR increases, symbol usage becomes more balanced.
In the noiseless limit, $\hat J=J$ and $I(J;\hat J)=H(J)$, so the uniform distribution maximizes mutual information.

\subsection{Relation to the learned symbol usage}

The reference distribution optimizes mutual information over freely chosen symbol probabilities.
In SemQAM, however, these probabilities are determined by the learned image representation and VQ assignments.
For a latent residual $u$ and VQ codewords $c_j$, the probability of transmitting index $j$ is
\begin{equation}
p_j
=
\Pr\left(
\|u-c_j\|^2
\leq
\|u-c_m\|^2,\ \forall m
\right).
\label{eq:learned_symbol_probability}
\end{equation}
The resulting distribution therefore depends on the latent features and codebook geometry.

The learned symbol usage shows a clear similarity to the reference in \cref{fig:information_theoretic_distribution}.
Both favor corner symbols at low SNR and become increasingly balanced as the channel improves.
This agreement suggests that SemQAM implicitly captures the influence of channel conditions on the preferred symbol distribution.

However, SemQAM uses a considerably denser distribution than the channel reference.
The reference maximizes preserved channel information, while SemQAM must also maintain sufficient codeword diversity for image representation.
Its generative reconstruction objective and autoregressive dependencies further affect which symbols are useful for image coding.
The learned distribution therefore reflects both channel characteristics and the requirements of image reconstruction.

Overall, SemQAM exhibits a close connection to the information-theoretic reference while learning a different distribution to accommodate generative reconstruction and autoregressive coding.
This comparison suggests that adapting symbol probabilities to the channel is an important source of the transmission efficiency gains achieved by joint coding over separated systems.

%% file: data/resolution_table.tex
\begin{table*}[t]\centering
\caption{High-resolution encoding and decoding latency (ms) and peak GPU memory (GiB) on one NVIDIA A100.
Bold and underline mark the best and second-best results, respectively.}
\label{tab:resolution}
\begin{tabular}{lrrrrrrrrr}
\toprule & \multicolumn{3}{c}{1080$\times$1920} & \multicolumn{3}{c}{1440$\times$2560} & \multicolumn{3}{c}{2160$\times$3840} \\
\cmidrule(lr){2-4}\cmidrule(lr){5-7}\cmidrule(lr){8-10}
Method & Enc. & Dec. & Mem. & Enc. & Dec. & Mem. & Enc. & Dec. & Mem. \\ \midrule
NTSCC \textcolor{black!60}{(JSAC'22)} & 130.97 & 138.18 & 7.24 & 245.81 & 371.34 & 12.95 & 1034.13 & 1368.10 & 28.55 \\
NTSCC+ \textcolor{black!60}{(JSTSP'23)} & 222.02 & 181.14 & 3.96 & 395.41 & 320.84 & 6.85 & 849.50 & 676.75 & 14.28 \\
MDJCM-A \textcolor{black!60}{(TCOM'25)} & 141.46 & 279.99 & 7.69 & 250.12 & 499.22 & 13.57 & 753.93 & 1103.39 & 28.69 \\
Quad-DeepSC \textcolor{black!60}{(JSAC'26)} & 87.09 & 127.82 & 3.56 & 155.72 & 225.57 & 6.27 & 353.91 & 505.65 & 13.65 \\
\midrule
MS-ILLM \textcolor{black!60}{(ICML'23)} & 351.13 & 395.19 & 2.99 & 516.88 & 629.61 & 4.87 & 1306.75 & 1677.35 & 9.94 \\
OSCAR \textcolor{black!60}{(NeurIPS'25)} & 513.39 & 1131.38 & 24.75 & 2382.80 & 2733.00 & 79.44 & \textemdash & \textemdash & $>80$ \\
OneDC \textcolor{black!60}{(NeurIPS'25)} & 303.50 & 748.65 & 12.81 & 585.94 & 1565.03 & 18.86 & 1542.53 & 4869.25 & 35.38 \\
DiT-IC \textcolor{black!60}{(CVPR'26)} & 480.42 & 758.49 & 10.98 & 718.43 & 1125.23 & 14.47 & 1582.35 & 2481.06 & 25.41 \\
EF-LIC \textcolor{black!60}{(ICML'26)} & \underline{19.79} & \underline{49.21} & \textbf{0.65} & \underline{31.96} & \underline{81.04} & \textbf{1.04} & \underline{65.99} & \underline{175.01} & \textbf{2.10} \\
CoD-Lite \textcolor{black!60}{(ICML'26)} & 42.65 & 77.34 & 2.45 & 74.03 & 133.54 & 3.97 & 164.25 & 297.66 & 8.33 \\
\midrule
SemQAM & \textbf{13.88} & \textbf{11.97} & \underline{0.711} & \textbf{22.76} & \textbf{20.15} & \underline{1.311} & \textbf{34.06} & \textbf{44.93} & \underline{2.332} \\
\bottomrule\end{tabular}\end{table*}

%% file: appendix/2_additional_experimental_results.tex
\section{Additional experimental results}
\label{app:additional_results}

\begin{figure*}[t]
\centering
\includegraphics[width=\linewidth]{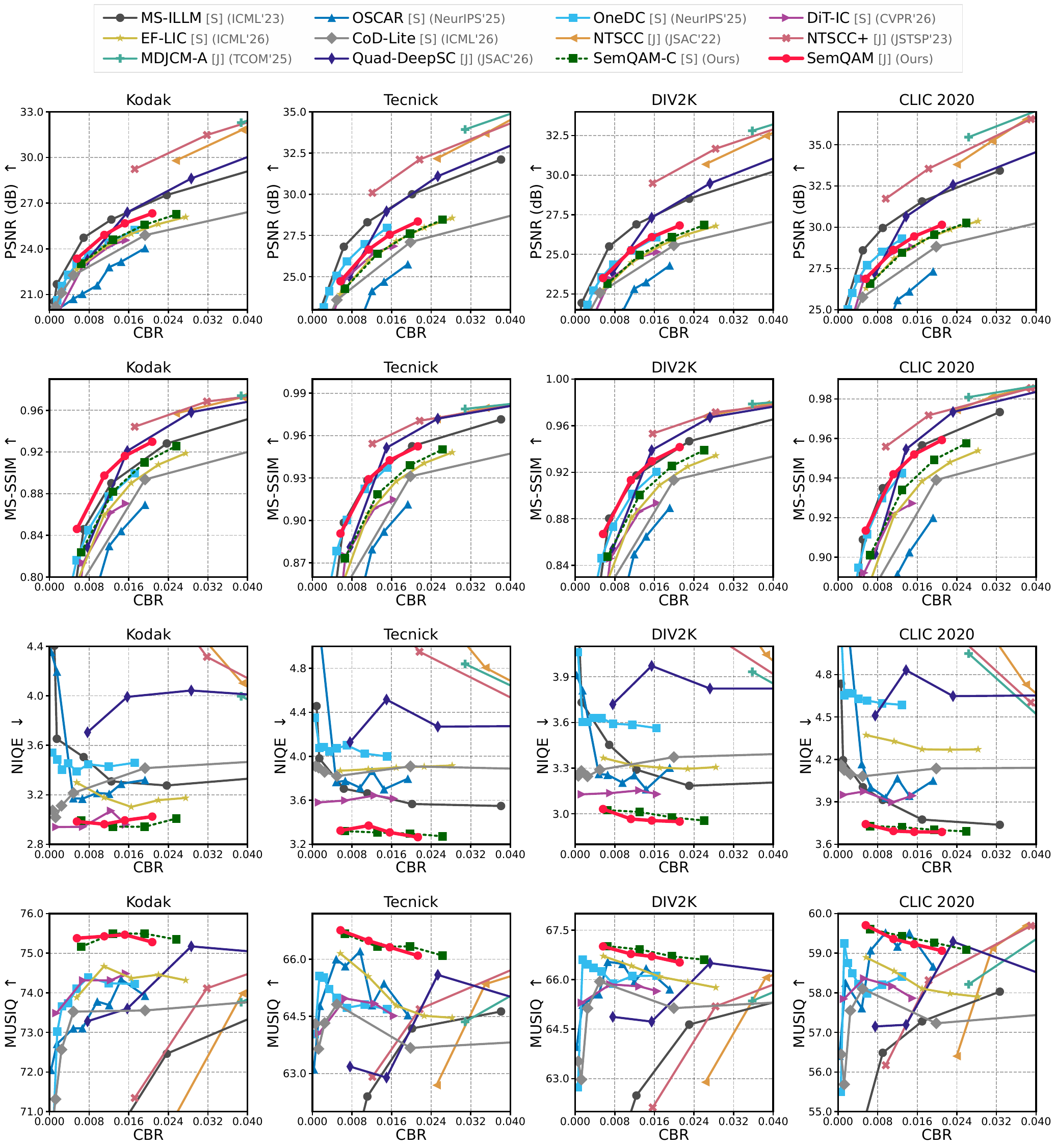}
\caption{Additional rate and reconstruction quality comparisons across four datasets over a 10 dB AWGN channel.
Higher PSNR, MS-SSIM, and MUSIQ and lower NIQE indicate better quality.}
\label{fig:app_10db_other_metrics}
\end{figure*}

\begin{figure*}[t]
\centering
\includegraphics[width=\linewidth]{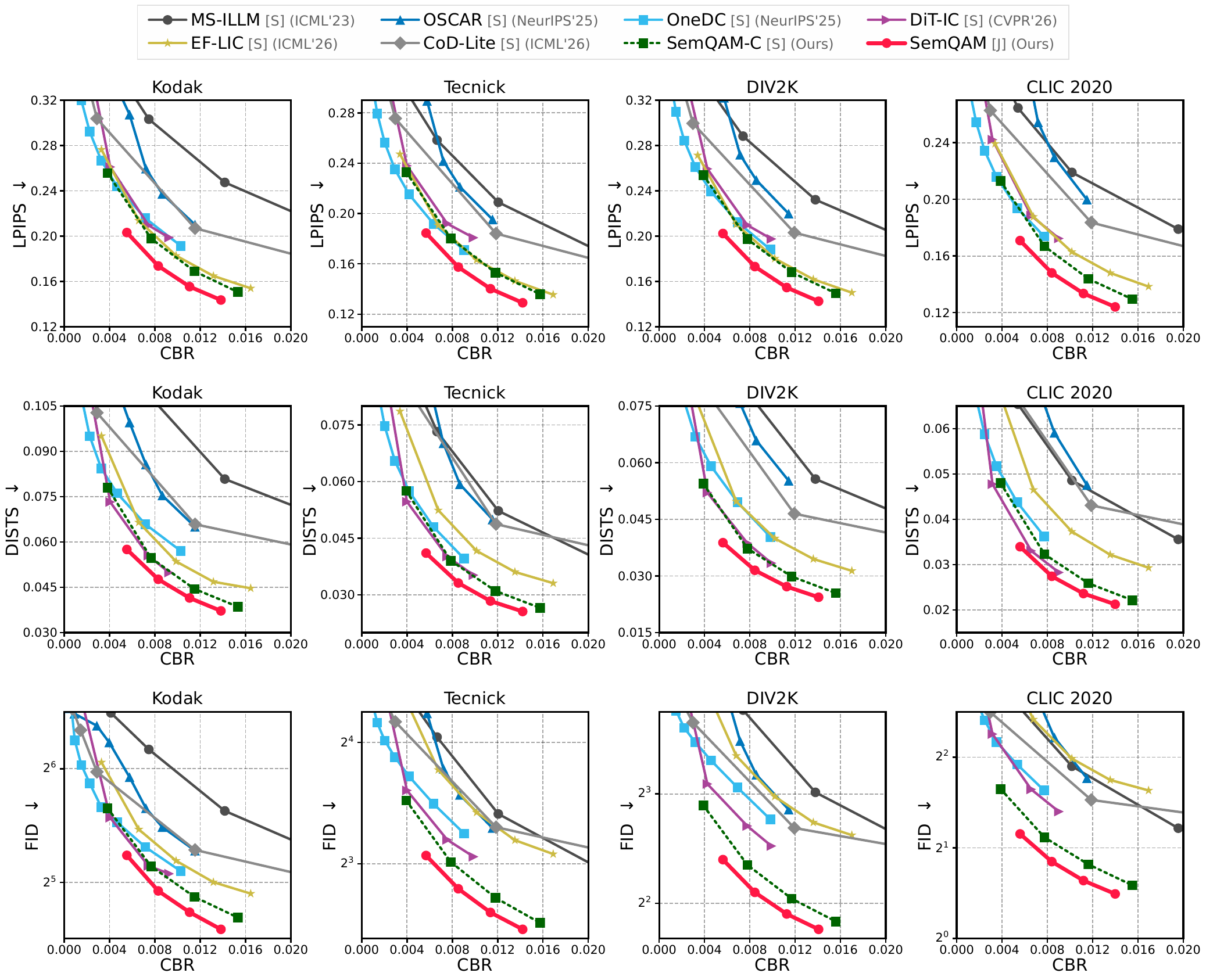}
\caption{Rate and perceptual quality comparisons across four datasets over a 16 dB AWGN channel.
Lower LPIPS, DISTS, and FID indicate better reconstruction quality.}
\label{fig:app16}
\end{figure*}

\begin{figure*}[t]
\centering
\includegraphics[width=\linewidth]{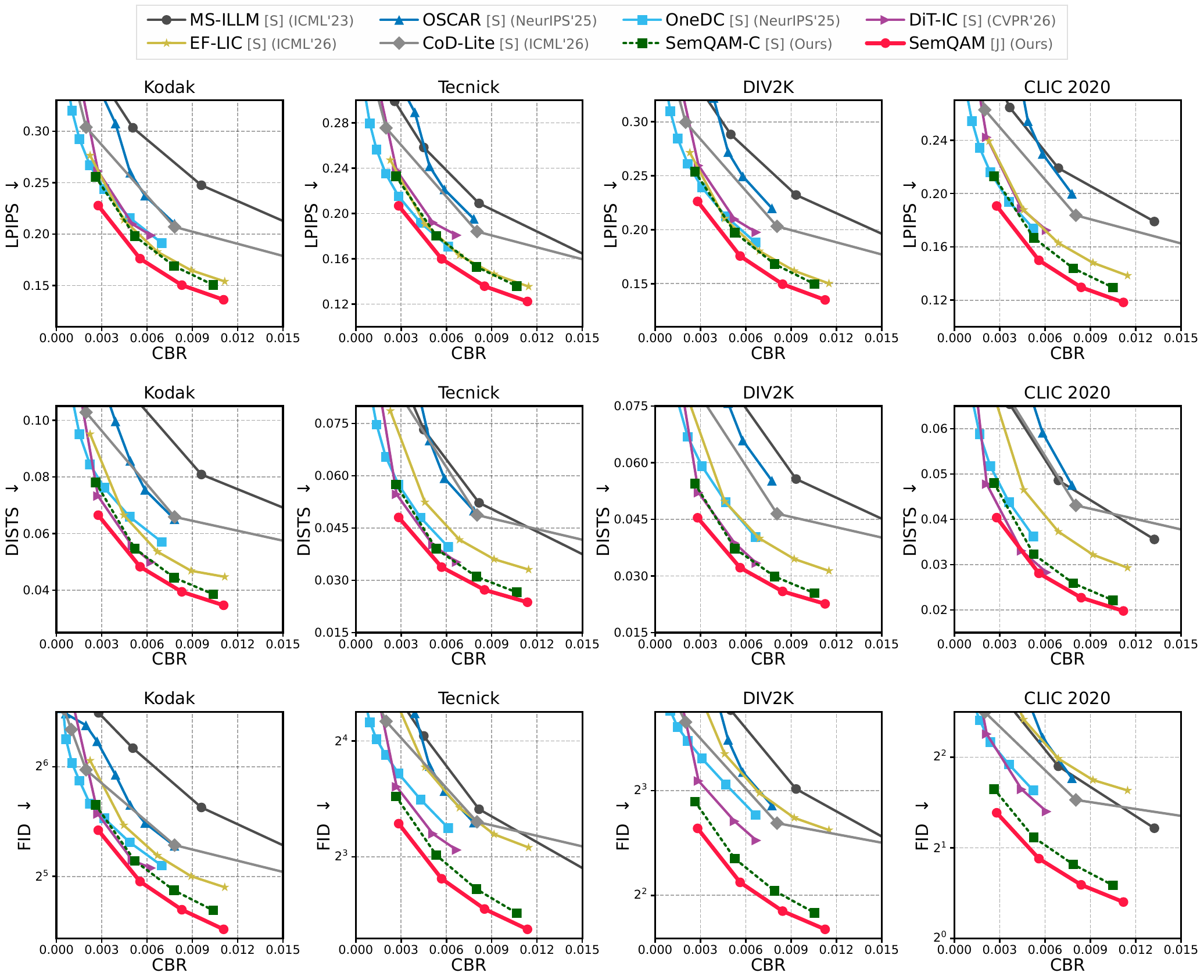}
\caption{Rate and perceptual quality comparisons across four datasets over a 24 dB AWGN channel.
Lower LPIPS, DISTS, and FID indicate better reconstruction quality.}
\label{fig:app24}
\end{figure*}

\begin{figure}[t]
    \centering
    \includegraphics[width=\columnwidth]{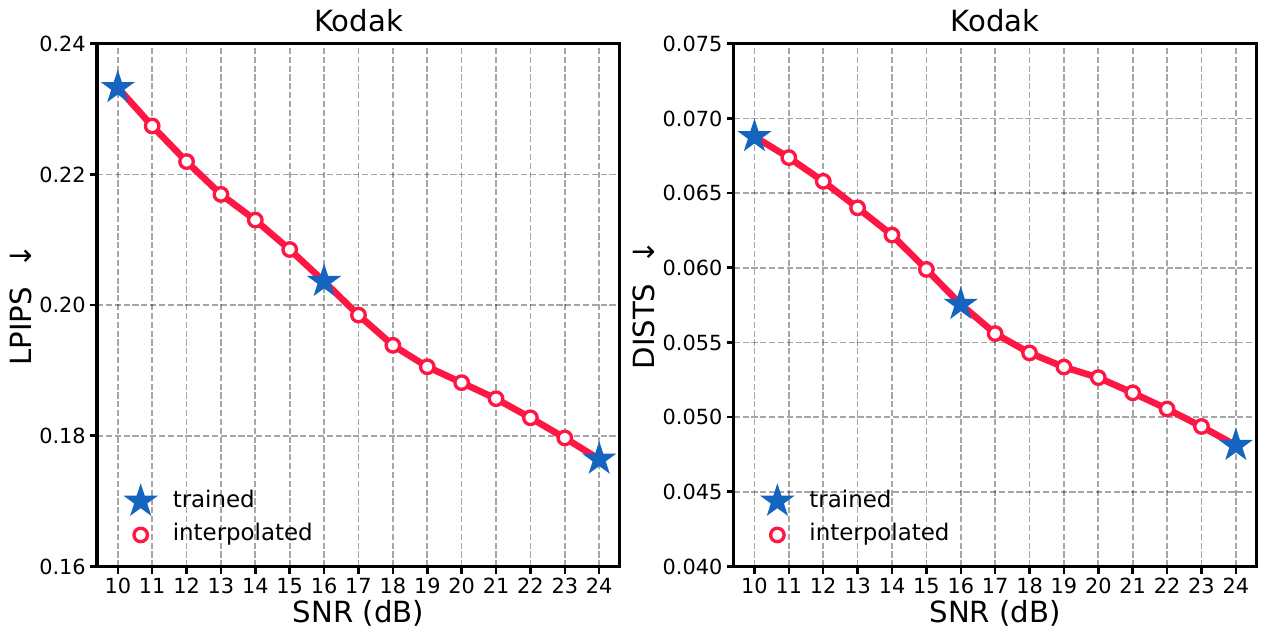}
    \caption{SNR interpolation on Kodak. Stars indicate trained settings, and circles indicate interpolated settings without retraining.}
    \label{fig:snr_interpolation}
    \vspace{-4.6pt}
\end{figure}

\input{data/all_bd_table}

This section extends the main experiments with additional quality metrics, comparisons at 16 and 24 dB, results across four datasets, and latency and memory measurements at higher resolutions.
We further examine the stability of learned symbol distributions and reconstruction quality across random initializations and channel realizations.

\subsection{Higher resolution execution}
\label{app:resolution}

To evaluate whether SemQAM maintains its efficiency as image resolution increases, we measure coding latency and peak GPU memory at 1080p, $1440\times2560$, and 4K on an NVIDIA A100.
As shown in \cref{tab:resolution}, SemQAM achieves the lowest encoding and decoding latency among the compared methods at all three resolutions.
At 4K, SemQAM reduces encoding and decoding latency by 48.4\% and 74.3\% compared with EF-LIC, with only 11.0\% higher peak GPU memory.
These results demonstrate that SemQAM retains its efficiency at high resolutions, highlighting its lightweight design and practical value.

\subsection{Additional quality metrics}

We further evaluate SemQAM and SemQAM-C using PSNR, MS-SSIM, NIQE, and MUSIQ on four datasets over a 10 dB AWGN channel, as shown in \cref{fig:app_10db_other_metrics}.
NTSCC, NTSCC+, and MDJCM-A are primarily optimized for PSNR and achieve strong pixel fidelity, but perform less favorably on perceptual metrics.
This observation is consistent with the tradeoff between distortion and perception~\citep{blau2018pd}, where higher PSNR does not necessarily indicate better perceptual quality.
Compared with other generative codecs, SemQAM and SemQAM-C achieve favorable NIQE and MUSIQ results while maintaining competitive PSNR and MS-SSIM.
These results further demonstrate the strong perceptual reconstruction capability of SemQAM in both joint transmission and conventional source coding.

\subsection{Additional channel conditions}

\paragraph{Performance at higher SNRs.}

To examine whether the gains from joint optimization extend to higher SNRs, we further evaluate SemQAM over 16 and 24 dB AWGN channels.
As shown in \cref{fig:app16,fig:app24}, SemQAM consistently outperforms the separated SemQAM-C system in LPIPS, DISTS, and FID across all four datasets.
These results demonstrate that adapting image representations to channel impairments improves transmission efficiency across different channel conditions while retaining standard 1024-QAM modulation.

\paragraph{Interpolation across SNRs.}

SemQAM also supports intermediate SNRs between its trained settings of 10, 16, and 24 dB without retraining.
For a target SNR, we select the two neighboring settings and retain the codebooks and index mappings from one of them.
We interpolate their cumulative RVQ outputs and compensate for the resulting energy reduction to construct the target response.
We then jointly adjust the output projections using least squares with a penalty based on channel noise variance.
The input projections are adjusted sequentially to compensate for changes in residuals.
We limit the correction magnitude to preserve the learned representations.

As shown in \cref{fig:snr_interpolation}, both LPIPS and DISTS improve smoothly across the 10--16 dB and 16--24 dB intervals at CBR = 0.00553.
These results demonstrate that SemQAM can adapt to intermediate SNRs without retraining.
In practice, the required configurations can be precomputed for expected SNRs and selected using channel quality estimates, avoiding optimization during transmission.
Extending coverage to other SNR ranges is an engineering task that depends on deployment requirements.

The compared JSCC methods also primarily evaluate transmission under specified channel conditions and known SNRs.
Conventional separated systems similarly rely on channel estimation to select suitable modulation and coding schemes.
Generalization without SNR information requires a different system model and lies beyond the practical setting and scope of this work.

\subsection{Performance across datasets}
\label{app:more_results}

To evaluate whether the gains from joint optimization extend across datasets, we compare SemQAM with previous JSCC and generative image coding methods on Kodak, Tecnick, DIV2K, and CLIC 2020.
As shown in \cref{tab:all_bd}, SemQAM-C achieves perceptual performance competitive with the recent DiT-IC, while substantially reducing coding latency and computational complexity.
It also consistently outperforms the lightweight EF-LIC and CoD-Lite across LPIPS, DISTS, and FID.
While retaining strong conventional source coding performance, SemQAM achieves the lowest BD-CBR across all three metrics on all four datasets, with approximately 30\% savings over SemQAM-C.
These results demonstrate the effectiveness of both conventional source coding and joint optimization across different image content and resolutions.

\subsection{Robustness to initialization and channel noise}
\label{app:seeds}

\paragraph{Initialization robustness.}
To examine the sensitivity of probability shaping to initialization, we train SemQAM with five independent random initializations of the network and VQ codebooks.
The resulting models exhibit similar QAM symbol usage patterns and comparable reconstruction performance.
These observations suggest that the learned shaping behavior reflects the channel transition structure rather than a particular initialization.

\paragraph{Robustness to channel noise.}
To evaluate reconstruction stability under random channel noise, we test SemQAM over ten independent AWGN realizations across all sixteen operating points on Kodak.
At 10 dB, the LPIPS standard deviations are only $0.000463$ and $0.000165$ at CBRs of $0.005534$ and $0.020752$, corresponding to approximately $0.20\%$ and $0.11\%$ of their respective means.
The corresponding PSNR standard deviations are $0.0219$ dB and $0.0087$ dB.
These small variations suggest that SemQAM learns representations robust to channel noise, maintaining consistent reconstruction quality across different noise realizations.

%% file: data/all_bd_table.tex
\begin{table*}[t]
\centering
\caption{BD-CBR comparison across four datasets over a 10 dB AWGN channel.
SemQAM-C is the anchor, and lower values are better.
Bold and underline mark the best and second-best results, respectively.
A dash denotes an unavailable comparison.}
\label{tab:all_bd}
\begin{tabular}{lllrrrr}
\toprule
Metric & Family & Method & Kodak & Tecnick & DIV2K & CLIC \\
\midrule
\multirow{12}{*}{LPIPS}
& \multirow{4}{*}{JSCC} & NTSCC \textcolor{black!60}{(JSAC'22)} & 359.10 & 359.87 & 334.58 & \textemdash \\
& & NTSCC+ \textcolor{black!60}{(JSTSP'23)} & 336.34 & 348.32 & 308.49 & 701.65 \\
& & MDJCM-A \textcolor{black!60}{(TCOM'25)} & 336.90 & 329.30 & 289.27 & 644.86 \\
& & Quad-DeepSC \textcolor{black!60}{(JSAC'26)} & 34.60 & 7.82 & 14.65 & 20.20 \\
\cmidrule(l){2-7}
& \multirow{6}{*}{GIC} & MS-ILLM \textcolor{black!60}{(ICML'23)} & 233.82 & 125.73 & 171.89 & 197.92 \\
& & OSCAR \textcolor{black!60}{(NeurIPS'25)} & 82.95 & 87.82 & 103.13 & \textemdash \\
& & OneDC \textcolor{black!60}{(NeurIPS'25)} & 13.84 & \underline{-9.97} & 4.19 & 4.80 \\
& & DiT-IC \textcolor{black!60}{(CVPR'26)} & 12.22 & 11.02 & 16.75 & 20.61 \\
& & EF-LIC \textcolor{black!60}{(ICML'26)} & 5.97 & 0.58 & 4.40 & 23.22 \\
& & CoD-Lite \textcolor{black!60}{(ICML'26)} & 67.24 & 57.11 & 63.11 & 103.31 \\
\cmidrule(l){2-7}
& \multirow{2}{*}{\textbf{SemQAM}} & SemQAM-C & \underline{0.00} & 0.00 & \underline{0.00} & \underline{0.00} \\
& & SemQAM & \textbf{-30.85} & \textbf{-31.88} & \textbf{-31.12} & \textbf{-31.23} \\
\midrule
\multirow{12}{*}{DISTS}
& \multirow{4}{*}{JSCC} & NTSCC \textcolor{black!60}{(JSAC'22)} & 696.09 & 742.78 & 839.13 & \textemdash \\
& & NTSCC+ \textcolor{black!60}{(JSTSP'23)} & 603.25 & 695.05 & 689.10 & \textemdash \\
& & MDJCM-A \textcolor{black!60}{(TCOM'25)} & 634.77 & 699.53 & 715.74 & \textemdash \\
& & Quad-DeepSC \textcolor{black!60}{(JSAC'26)} & 89.99 & 65.09 & 74.37 & 66.34 \\
\cmidrule(l){2-7}
& \multirow{6}{*}{GIC} & MS-ILLM \textcolor{black!60}{(ICML'23)} & 324.82 & 159.08 & 278.00 & 186.48 \\
& & OSCAR \textcolor{black!60}{(NeurIPS'25)} & 111.56 & 129.00 & \textemdash & 192.97 \\
& & OneDC \textcolor{black!60}{(NeurIPS'25)} & 30.88 & 14.77 & 48.79 & 19.27 \\
& & DiT-IC \textcolor{black!60}{(CVPR'26)} & \underline{-2.96} & \underline{-2.61} & 4.38 & \underline{-13.93} \\
& & EF-LIC \textcolor{black!60}{(ICML'26)} & 25.72 & 46.64 & 49.67 & 70.27 \\
& & CoD-Lite \textcolor{black!60}{(ICML'26)} & 139.64 & 137.75 & 147.97 & 196.31 \\
\cmidrule(l){2-7}
& \multirow{2}{*}{\textbf{SemQAM}} & SemQAM-C & 0.00 & 0.00 & \underline{0.00} & 0.00 \\
& & SemQAM & \textbf{-28.06} & \textbf{-29.57} & \textbf{-30.25} & \textbf{-29.78} \\
\midrule
\multirow{12}{*}{FID}
& \multirow{4}{*}{JSCC} & NTSCC \textcolor{black!60}{(JSAC'22)} & 494.21 & \textemdash & \textemdash & \textemdash \\
& & NTSCC+ \textcolor{black!60}{(JSTSP'23)} & 442.60 & 913.92 & 1112.09 & \textemdash \\
& & MDJCM-A \textcolor{black!60}{(TCOM'25)} & 434.92 & 861.29 & 1016.81 & \textemdash \\
& & Quad-DeepSC \textcolor{black!60}{(JSAC'26)} & 98.34 & 164.50 & 214.39 & 287.10 \\
\cmidrule(l){2-7}
& \multirow{6}{*}{GIC} & MS-ILLM \textcolor{black!60}{(ICML'23)} & 245.39 & 154.50 & 269.92 & 216.04 \\
& & OSCAR \textcolor{black!60}{(NeurIPS'25)} & 84.34 & 123.54 & \textemdash & \textemdash \\
& & OneDC \textcolor{black!60}{(NeurIPS'25)} & 11.38 & 58.42 & 118.59 & 95.54 \\
& & DiT-IC \textcolor{black!60}{(CVPR'26)} & \underline{-4.77} & 20.24 & 56.35 & 69.90 \\
& & EF-LIC \textcolor{black!60}{(ICML'26)} & 37.50 & 124.57 & 191.79 & 325.95 \\
& & CoD-Lite \textcolor{black!60}{(ICML'26)} & 86.00 & 151.56 & 177.82 & 221.82 \\
\cmidrule(l){2-7}
& \multirow{2}{*}{\textbf{SemQAM}} & SemQAM-C & 0.00 & \underline{0.00} & \underline{0.00} & \underline{0.00} \\
& & SemQAM & \textbf{-28.80} & \textbf{-29.07} & \textbf{-31.83} & \textbf{-33.85} \\
\bottomrule
\end{tabular}
\end{table*}

%% file: appendix/3_additional_method_details.tex
\section{Additional method details}
\label{app:method_details}

We provide additional details of the SemQAM architecture, analyze how channel transitions affect codebook learning at the receiver, and explain why we use 1024-QAM.

\begin{figure*}[t]
    \centering
    \includegraphics[width=\linewidth]{figures/semqam_arch.pdf}
    \caption{Architecture of SemQAM.
    The image transforms and context networks are shared across operating points, while each operating point uses its own RVQ codebooks.}
    \label{fig:app_framework}
\end{figure*}

\subsection{Model architecture}
\label{app:architecture}

As illustrated in \cref{fig:app_framework}, SemQAM retains the image transforms, hyperprior, quadtree context model, and projected RVQ of EF-LIC~\citep{cao2026eflic}.
Each operating point uses its own RVQ codebooks containing 1024 codewords, with 16-dimensional vectors for the main latent and 8-dimensional vectors for the hyperlatent.

Unlike EF-LIC, SemQAM introduces a multiplicative normalization field $q$ following DCVC-RT~\citep{jia2025dcvcrt}.
The hyper synthesis transform $h_s$ predicts this field from the reconstructed hyperlatent, with its values constrained to be at least $0.5$.
The main latent $y$ is divided by $q$ before quantization, and the reconstructed latent is multiplied by $q$ before image synthesis.
This provides additional control over the latent scale.

For quadtree group $i$, let $y_i$ denote the scaled latent group.
The context features $\psi_i$ combine the hyperprior features $\phi$ with previously reconstructed groups $\hat y_{<i}$.
The predictor $f_i^{\mathrm{RD}}$ produces the mean $\mu_i$ and scale $\sigma_i$ used by the affine transforms.

Before RVQ, the encoder transform $e_i^{\mathrm{RD}}$ removes the predicted mean and normalizes the latent:
\begin{equation}
y_i'
=
\sigma_i^{-1}\odot(y_i-\mu_i).
\end{equation}
After RVQ reconstructs $\hat y_i'$, the decoder transform $d_i^{\mathrm{RD}}$ restores the corresponding group:
\begin{equation}
\hat y_i
=
\sigma_i\odot\hat y_i'+\mu_i.
\end{equation}
Here, $\odot$ denotes elementwise multiplication.
The reconstructed groups are merged and rescaled by $q$ before the synthesis transform $g_s$.

For clarity, \cref{fig:app_framework} illustrates a shared coding architecture without separately depicting the transmitter and receiver pathways.
In practice, the transmitter uses estimated receiver reconstructions based on Expected-RX, while the receiver uses the actual reconstructions obtained from detected indices.

\subsection{Receiver-side codebook learning}
\label{app:receiver_loss_equivalence}

We analyze how channel transitions influence the learned codebook.
For one quantization step, let $J$ and $\hat J$ denote the transmitted and detected indices, $c_k$ the $k$-th codeword, and $\hat u$ the receiver-side latent target.

Let $p_j=\Pr(J=j)$ denote the probability of transmitting index $j$, and let $W_{kj}=\Pr(\hat J=k\mid J=j)$ denote the channel transition probability.
The expected receiver-side codebook loss is
\begin{equation}
\mathbb E[\mathcal L_{\rm cb}]=\sum_{j,k}p_jW_{kj}\mathbb E\left[\|\operatorname{sg}(\hat u)-c_k\|^2\mid J=j,\hat J=k\right],
\label{eq:expected_rx_codebook_loss_full}
\end{equation}
where $\operatorname{sg}$ denotes stop-gradient.
Thus, training with detected indices incorporates channel transition probabilities into codebook optimization.

For fixed assignments and detached targets, the minimizing codeword for a detected index $k$ with $\Pr(\hat J=k)>0$ is
\begin{equation}
c_k^*=\mathbb E[\hat u\mid\hat J=k]=\frac{\sum_j p_jW_{kj}\mathbb E[\hat u\mid J=j,\hat J=k]}{\sum_j p_jW_{kj}}.
\label{eq:receiver_centroid}
\end{equation}

This result shows that each receiver codeword is influenced by latent targets associated with different transmitted indices, weighted by their usage probabilities and channel transitions.
Frequently confused QAM symbols therefore receive overlapping training signals, encouraging compatible latent representations.
These updates reshape the codebook geometry and its nearest-neighbor assignments, allowing symbol usage to adapt to channel conditions without an explicit probability shaping objective.

\subsection{Choice of modulation order}
\label{app:modulation_order}

Higher QAM orders allow each channel symbol to represent more bits, potentially improving transmission efficiency.
However, higher orders are more sensitive to noise and require better channel conditions for reliable detection.
Separated systems therefore typically select lower modulation orders at low SNR to ensure reliable LDPC decoding.
For example, 16-QAM can be more suitable than 1024-QAM over a 10 dB AWGN channel, depending on the coding rate and target reliability.

SemQAM addresses this limitation through joint optimization of image representations and QAM symbols.
By learning codewords that account for channel detection errors, SemQAM can tolerate symbol errors without requiring exact recovery of every transmitted index.
Our experiments demonstrate that SemQAM achieves strong reconstruction quality using fixed 1024-QAM across 10, 16, and 24 dB AWGN channels, as well as a Rayleigh fading channel.
Thus, SemQAM does not require modulation order switching under the evaluated channel conditions.

We choose 1024-QAM because it provides a large symbol alphabet while retaining practical hardware support.
It is supported by Wi-Fi 6~\citep{ieee80211ax2021,intelwifi6} and was introduced for 5G NR downlink transmission in Release 17~\citep{threegpp38211rel17}.
Although the newer Wi-Fi 7 standard supports 4096-QAM~\citep{ieee80211be2024}, its hardware deployment is still expanding.
We therefore adopt 1024-QAM to balance the constellation size with hardware compatibility.
Moreover, its 1024 constellation points directly match the 1024 entries in each VQ codebook, enabling a simple one-to-one mapping without additional bit conversion.

%% file: appendix/4_training_evaluation_protocol.tex
\section{Detailed evaluation protocol}
\label{app:protocol}

This section describes the evaluation protocols used in our experiments.
We first explain the retransmission protocol and MCS selection for separated systems.
We then describe baseline inference settings, channel conditions, image quality evaluation, and metric computation.

\subsection{Separated transmission evaluation}
\label{app:separated_transmission}

Unlike JSCC methods that reconstruct images directly from received channel symbols, separated systems require correct recovery of the compressed bitstream.
Errors remaining after LDPC decoding may prevent image reconstruction and cause severe quality degradation.
Following Quad-DeepSC~\citep{huang2026quaddeepsc}, we retransmit failed transmissions and evaluate the first successful reconstruction.
All retransmissions are included in CBR.

We further search the 3GPP modulation and coding scheme (MCS) table and select the configuration with the lowest measured CBR, including retransmission costs.
This provides the best transmission efficiency among the evaluated MCS configurations and avoids penalizing separated systems with corrupted bitstreams.
Even under this favorable evaluation protocol, SemQAM achieves substantial BD-CBR savings while reconstructing images without retransmission.

\subsection{Baseline inference protocol}
\label{app:baselines}

\paragraph{Learned JSCC and JSCM methods.}

This section describes how we evaluate existing joint coding methods and explains the choice of comparison settings.
We use the official implementations and model weights provided by the original authors.
For Quad-DeepSC~\citep{huang2026quaddeepsc}, we contacted the authors and obtained its source code and a model optimized for LPIPS, enabling a fair comparison under perceptual reconstruction objectives.

Although MDJCM~\citep{zhang2024mdjcm} proposes joint coding and modulation with finite constellations, only its analog variant MDJCM-A is publicly available.
MDJCM-A transmits continuous channel symbols, providing an idealized performance reference without the constraints of finite QAM constellations.
We therefore include MDJCM-A as a strong baseline for comparison.

NTSCC+~\citep{wang2023ntsccplus} and Quad-DeepSC require separate reliable transmission of certain auxiliary information, whereas SemQAM directly transmits its VQ indices through standard QAM without separate channel coding.
All required transmission resources are included in CBR to ensure a fair comparison.

Unlike SemQAM, the final Quad-DeepSC model~\citep{huang2026quaddeepsc} does not support conventional source coding.
Although Quad-DeepSC first trains a source codec called Quad-LIC to facilitate JSCC training, it subsequently updates the pretrained parameters for joint transmission.
The resulting model transmits continuous channel symbols and does not provide quantization and entropy coding to generate a compressed binary bitstream.
In contrast, SemQAM supports both conventional source coding and joint transmission using the same trained model.

DeepJSCC-Q~\citep{tung2022deepjsccq} and VQ-DeepSC~\citep{fu2023vqdeepsc} are excluded from our main comparison because official implementations and pretrained models were unavailable.
Our preliminary reproductions showed substantially weaker performance than SemQAM, making further comparisons of limited value.
Recent methods such as MDJCM and Quad-DeepSC also omit these models from their reported comparisons.

CAVQ~\citep{meng2025cavq} also lacks an available official implementation.
We instead reproduce its core quantization strategy under the same experimental conditions as SemQAM.
This configuration is included as $M_0$ in our ablation study, where the complete SemQAM achieves substantially better performance.

\paragraph{Separated generative codecs.}

We evaluate separated generative codecs using their official implementations, released checkpoints, and inference procedures.
For CoD-Lite, we use the released 0.0312 bpp checkpoint.
At $1024\times1920$, the official implementation takes 39.52 ms for encoding and 56.86 ms for decoding on our NVIDIA A100, compared with 16 ms and 24 ms reported in the original paper.
SemQAM takes only 13.88 ms and 11.97 ms at the slightly higher resolution of $1080\times1920$, outperforming even the published CoD-Lite latency.
For consistent comparisons, we report measurements from the released implementations under our common hardware and timing protocol.

\subsection{Evaluation across channel conditions}
\label{app:snr_choice}

We use 10 dB as the primary channel condition for comparing learned JSCC methods.
This setting is widely used in image JSCC evaluations, including DeepJSCC-f, DeepJSCC-Q, NTSCC, NTSCC+, SwinJSCC, MDJCM, and Quad-DeepSC~\citep{kurka2020deepjsccf,tung2022deepjsccq,dai2022ntscc,wang2023ntsccplus,yang2025swinjscc,zhang2024mdjcm,huang2026quaddeepsc}.
Moreover, all learned JSCC baselines in our main comparison have official or author-provided models at 10 dB, while models at other SNRs are unavailable.
Evaluating these methods at different SNRs would require retraining and deviate from their released models.
We therefore restrict comparisons with learned JSCC methods to 10 dB.
At 16 and 24 dB, we report comparisons with separated systems only.

Our evaluation follows the common assumption that channel conditions can be estimated.
The compared JSCC methods primarily evaluate performance under specified SNRs.
Conventional separated systems also rely on channel state information (CSI) to select suitable modulation and coding schemes.
Therefore, transmission with estimated channel information is a common and practical setting, which SemQAM is designed to address.
Although SemQAM is trained at discrete SNRs, our interpolation results demonstrate its ability to generalize to intermediate SNRs without retraining.
This extends its applicability beyond the original training conditions.
Further generalization without any SNR estimate requires a different system model and is beyond the scope of this work.

\subsection{Image quality evaluation}
\label{app:evaluations}

We use objective image quality metrics as the primary quantitative evaluation, together with visual comparisons.
We do not conduct subjective user studies because human ratings can vary substantially across participants and viewing conditions, making the results difficult to reproduce.
We therefore report multiple metrics to provide a broad evaluation of reconstruction quality.

LPIPS with VGG features~\citep{zhang2018lpips,simonyan2015vgg} and DISTS~\citep{ding2022dists} measure perceptual similarity to reference images, while FID measures the similarity between image feature distributions.
NIQE and MUSIQ provide additional quality measures without reference images.
PSNR and MS-SSIM measure pixel distortion and structural similarity.

The perception--distortion tradeoff shows that improving perceptual quality can increase pixel distortion. Its extension to lossy compression further establishes a tradeoff among rate, distortion, and perception~\citep{blau2018pd,blau2019rdp}.
Since SemQAM targets perceptual reconstruction, LPIPS, DISTS, and FID are our primary metrics, while NIQE, MUSIQ, PSNR, and MS-SSIM provide complementary evaluations.

\subsection{Metric computation}
\label{app:metric_computation}

\paragraph{FID.}

We follow the FID evaluation protocol of GLC~\citep{jia2024glc}, which is also adopted by many generative image compression methods~\citep{xue2025onedc,shi2026ditic}.
We use $64\times64$ patches for Kodak and $256\times256$ patches for Tecnick, DIV2K, and CLIC 2020.
Patches from all images in each dataset are pooled to estimate the Inception feature distributions.
This provides more samples for distribution estimation, particularly for relatively small test sets.

\paragraph{BD-CBR.}

In JSCC and JSCM, the channel input consists of transmitted symbols rather than bits.
We therefore use CBR to measure channel usage and extend the Bjøntegaard method~\citep{bjontegaard2001} from bitrate--quality curves to CBR--quality curves~\citep{huang2026quaddeepsc}.
For each comparison, one method is selected as the anchor and assigned a BD-CBR of $0.00\%$.
We interpolate $\log_{10}(\mathrm{CBR})$ as a function of the original quality metric using piecewise cubic Hermite interpolation (PCHIP) over the common quality interval.
Let $a(q)$ and $b(q)$ denote the interpolated log-CBR curves of the anchor and comparison method over $[q_0,q_1]$.
The BD-CBR is computed as
\begin{equation}
\operatorname{BD\text{-}CBR}
=
100\left[
10^{
\frac{1}{q_1-q_0}
\int_{q_0}^{q_1}
(b(q)-a(q))\,dq
}
-1
\right].
\end{equation}
A negative value indicates fewer channel uses than the anchor at matched quality, while a positive value indicates more.
The main comparisons use SemQAM-C with LDPC and QAM as the anchor.
Only measured points within a valid common quality interval are used, and comparisons without such an interval are reported as unavailable.

\paragraph{BD-rate.}

For comparisons between conventional source codecs, we use the standard BD-rate, which measures bitrate differences at matched quality.
It follows the same interpolation and integration procedure, with bitrate replacing CBR.
Throughout the paper, BD-rate is used for source coding comparisons, while BD-CBR is used for joint transmission comparisons.

%% file: appendix/5_visual_comparison.tex
\section{Additional visual comparisons}
\label{app:visual_comparison}

\begin{figure*}[tp]
    \centering
    \includegraphics[width=0.8225\linewidth]{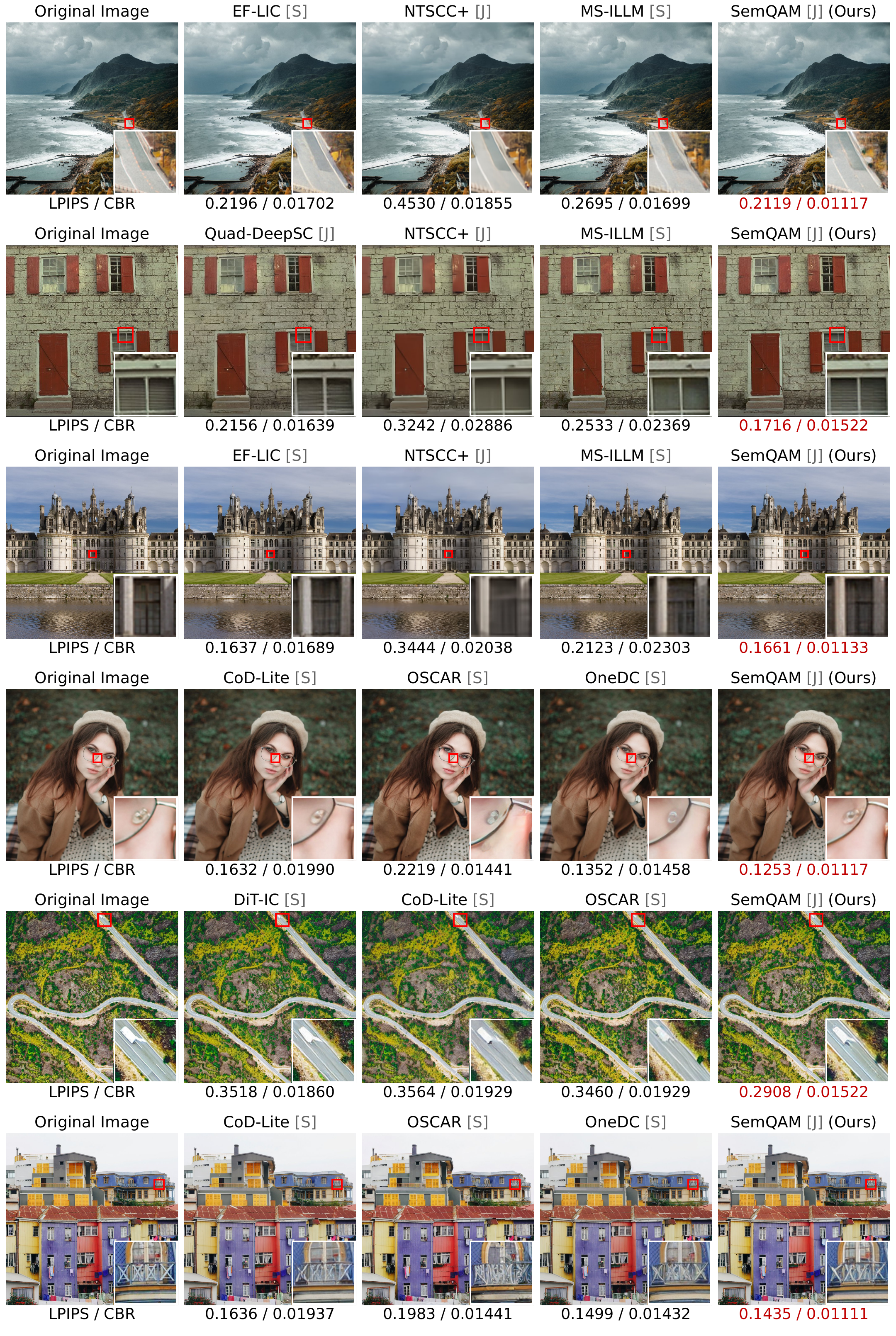}
    \caption{Visual comparisons on Kodak over a 10 dB AWGN channel.
    Insets highlight local details.
    Values report LPIPS and CBR, respectively, where lower is better.}
    \label{fig:visual_suppl}
\end{figure*}

\Cref{fig:visual_suppl} provides additional visual comparisons with existing methods.
SemQAM achieves the lowest CBR in all six examples and the lowest LPIPS in five of them.
The enlarged regions highlight preserved details, including window structures, the nose pad of the glasses, road markings, and railing patterns.
For the castle image, EF-LIC achieves slightly lower LPIPS (0.1637 versus 0.1661), while SemQAM requires fewer channel uses (0.01133 versus 0.01689 CBR).

